%% file: neurips_2026.tex
\documentclass{article}

\PassOptionsToPackage{numbers, compress}{natbib}

\usepackage[main, final]{neurips_2026}

\usepackage[utf8]{inputenc} 
\usepackage[T1]{fontenc}    
\usepackage{hyperref}       
\usepackage{url}            
\usepackage{booktabs}       
\usepackage{amsfonts}       
\usepackage{nicefrac}       
\usepackage{microtype}      
\usepackage{xcolor}
\usepackage{fancyvrb}

\definecolor{histblue}{RGB}{30,90,180}
\definecolor{midred}{RGB}{190,55,50}
\definecolor{markorange}{RGB}{210,105,30}

\newcommand{\histtok}[1]{\textcolor{histblue}{\texttt{#1}}}
\newcommand{\midtok}[1]{\textcolor{midred}{\texttt{#1}}}
\newcommand{\tok}[1]{\textcolor{markorange}{\texttt{#1}}}

\DefineVerbatimEnvironment{TokenTree}{Verbatim}{
  commandchars=@\{\},
  fontsize=\small,
  formatcom=\ttfamily
}

\usepackage[inline]{enumitem} 
\usepackage{url}
\usepackage{soul}
\usepackage{changes}
\usepackage{wrapfig}
\usepackage[table]{xcolor}
\definecolor{holdoutblue}{RGB}{235,244,252}
\definecolor{mtgreen}{RGB}{237,248,237}
\definecolor{intersectorange}{RGB}{252,241,227}

\usepackage{multirow} 

\usepackage{amsmath}   
\usepackage{amsthm}    
\usepackage{amsfonts}  
\usepackage{subfig}       
\usepackage{graphicx}

\usepackage{booktabs}
\usepackage{tabularx}

\definecolor{myblue}{RGB}{255,0,0}
\theoremstyle{definition}

\theoremstyle{plain}
\usepackage{algorithm}
\usepackage{algpseudocode}

\usepackage{float}
\title{TreeGraft: Adaptive Multi-Drafter Grafting for Tree-Based Speculative Decoding}

\author{%
  \textbf{Jiaming Fan}$^{1,2,3\ast}$ \quad
  \textbf{Daming Cao}$^{4\ast\ddagger}$ \quad
  \textbf{Canchen Huang}$^{1,2}$ \quad
  \textbf{Jiale Fu}$^{1,2}$ \quad
  \textbf{Jin Zhang}$^{3}$ \\
  \textbf{Junjie Gao}$^{3}$ \quad
  \textbf{Kai Yang}$^{3\dagger}$ \quad
  \textbf{Xiangzhong Luo}$^{1,2}$ \quad
  \textbf{Xu Yang}$^{1,2}$ \\
  $^{1}$Key Laboratory of New Generation Artificial Intelligence Technology and \\
  Its Interdisciplinary Applications (Southeast University), Ministry of Education, China \\
  $^{2}$Southeast University \quad $^{3}$Ant Group \\
  $^{4}$Nanjing University of Information Science and Technology \\
  {\small
    $^\ast$Equal Contribution \quad
    $^\dagger$Project Leader \quad
    $^\ddagger$Corresponding Author
  } \\
  {\small \texttt{jiaming.fan@seu.edu.cn} \quad \texttt{dmcao@nuist.edu.cn}} \\
  {\small \texttt{chengyue.yk@antgroup.com} \quad \texttt{xuyang\_palm@seu.edu.cn}}
}

\begin{document}

\maketitle

\begin{abstract}

Speculative decoding accelerates large language model inference through a draft-then-verify paradigm. Building on this, tree-structured methods improve inference by organizing proposals into multiple candidate paths, increasing the accepted length. However, existing tree-structured methods use a single drafter for all drafting steps, creating a dilemma: a smaller drafter is fast but yields lower-quality trees, whereas a larger drafter improves tree quality but suffers from high latency. To address this, we propose \textit{TreeGraft}, a multi-drafter framework in which drafters of different costs jointly construct a shared draft tree. TreeGraft uses the stronger drafter to rescore candidates by updating scores assigned by the weaker drafter, reselect grafting positions, and recover promising paths left unexplored. It also integrates stronger drafter expansions non-destructively, preserving existing branches that may still be accepted by the target model. Together, these designs improve the quality of the shared draft tree. To control the drafting cost, TreeGraft introduces a lightweight scheduler distilled from an offline value system to decide when to call the stronger drafter. Across 10 model pairs and 6 benchmarks, TreeGraft outperforms the better of the two fixed single-drafter endpoint strategies by 15.1\% on average, reaching a maximum gain of 26.6\%. Our code is available at \url{https://github.com/fjm9933/TreeGraft}.

\end{abstract}

\input{sections/sec-1}
\input{sections/sec-2}
\input{sections/sec-3}
\input{sections/sec-4}
\input{sections/sec-5}
\input{sections/sec-6}

\bibliographystyle{unsrtnat}
\bibliography{neurips_2026}

\newpage

\input{sections/appendix}

\newpage

\input{sections/checklist}

\end{document}

%% file: sections/sec-1.tex
\section{Introduction}

Autoregressive large language models (LLMs) achieve strong performance~\citep{https://doi.org/10.48550/arxiv.2503.07536,ICLR2026_b0d4b9d3,https://doi.org/10.48550/arxiv.2609.31140} across a wide range of tasks \citep{achiam2023gpt,guo2025deepseek,touvron2023llama,Liu_2026,https://doi.org/10.48550/arxiv.2609.37402,pmlr-v306-fu26j,NEURIPS2025_c2cd7fa9,https://doi.org/10.48550/arxiv.2509.23094,pmlr-v267-fu25c,Chen_2025,Jiang_2025_CVPR,NEURIPS2024_12d3e63b,NEURIPS2024_b619cd6d,Li_2024_CVPR}, yet their token-by-token decoding makes inference inherently slow. Early speculative decoding (SD)~\citep{leviathan23, chen23} accelerates this process by using a smaller model as the drafter, which constructs drafts that the target model then verifies in parallel. However, since the speedup is bounded by the accepted length, early SD faces a structural limitation: it restricts proposals to a single sequence. Consequently, once the target model rejects a token, all subsequent tokens are invalidated, limiting the accepted length. Tree-based SD \citep{Miao2023SpecInferAL,Li2024EAGLE2FI,Chen2024SequoiaSA} overcomes this by organizing proposals into a draft tree of multiple paths, utilizing alternative branches to bypass local rejections and thereby extend the accepted length. A long accepted length, however, requires a high-quality tree whose paths cover what the target model would accept, and producing such a tree incurs a substantial drafting cost. The central objective is therefore to build a high-quality tree at a low drafting cost.

\begin{figure}[!t]
    \centering
    \includegraphics[width=\linewidth]{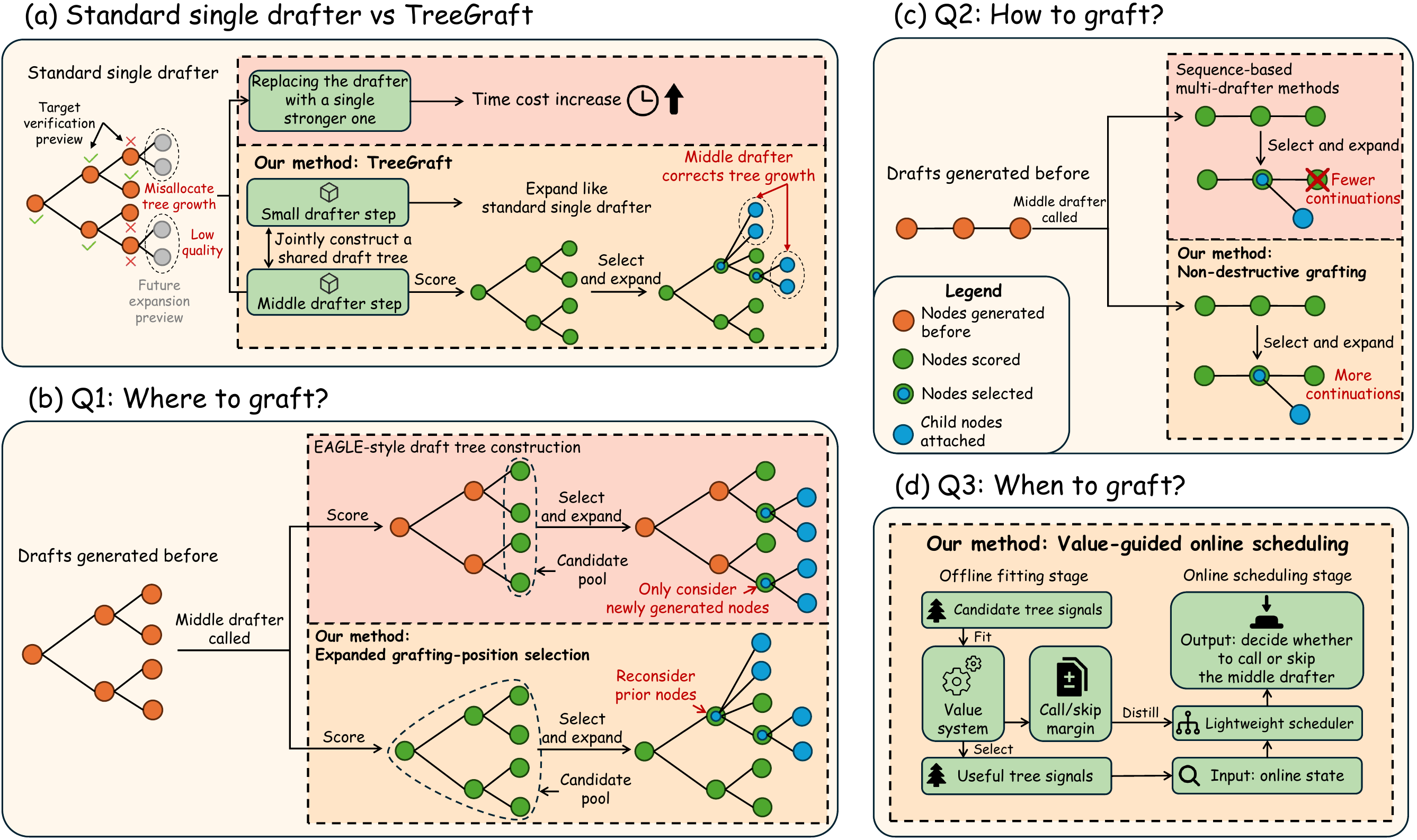}
    \caption{\textbf{Overview of TreeGraft.}
    (a) Standard single-drafter expansion is contrasted with TreeGraft. (b--d) The resulting design is organized around three questions: where to graft, how to graft, and when to graft.}
    \label{fig:motivation}
\end{figure}

To achieve this objective, we first examine the draft tree construction process. A general paradigm \citep{Li2024EAGLE2FI} is to build it incrementally over a series of drafting steps. At each step, the drafter forms a candidate pool from the current tree—typically the newly generated nodes from the previous step—and ranks these nodes according to a scoring criterion. The highest-ranked nodes are then expanded with new child nodes. However, existing tree-based methods mostly rely on a single drafter across all drafting steps~\citep{Miao2023SpecInferAL,Li2024EAGLE2FI}, which forces a binary trade-off between tree quality and per-step cost: a drafter with few parameters yields a lower-quality tree, while simply switching to a larger one raises tree quality but incurs higher latency at every step (Figure~\ref{fig:motivation}(a)). This raises a natural question: \textit{must the same drafter be used at every drafting step}? If a stronger drafter were invoked only at the steps where its quality gain matters most, the tree could be improved without paying its cost at all steps.

Motivated by this observation, we introduce two drafters of different costs---a lightweight small drafter and a stronger middle drafter---that jointly build a single draft tree, with each drafter handling a subset of drafting steps. We refer to this scheme as \textit{grafting}: nodes produced by one drafter can be added beneath nodes produced by the other on a shared draft tree. Making grafting work requires answering three questions. Within a single step, \textit{where} on the tree should new nodes be attached, and \textit{how} should they be combined with the existing tree? Across steps, \textit{when} should the middle drafter be called? The first two questions concern how grafting improves tree quality, and the third concerns how to control the drafting cost incurred by middle drafter calls.

\textbf{Where to graft.} In common single-drafter tree construction \citep{Li2024EAGLE2FI}, the candidate pool is restricted to the leaf nodes generated in the previous step. For a single drafter, this rule is natural: earlier candidates have already been evaluated by the same model, so tree construction advances from the newly generated frontier. However, directly applying this to a multi-drafter setting restricts the stronger middle drafter to the newest nodes, forcing it to blindly trust the existing scores on the shared tree. If a promising node was earlier under-ranked by the small drafter, that node is permanently ignored. We therefore expand the candidate pool beyond the previous step at middle drafter steps, allowing the middle drafter to rescore earlier nodes, reselect grafting positions, and therefore reopen prematurely ignored paths (Figure~\ref{fig:motivation}(b)).

\textbf{How to graft.} Existing hierarchical multi-drafter methods are largely sequence-based \citep{Chen2023CascadeSD, Sun2024TriForceLA}: when a stronger drafter generates new children under a parent, the parent's existing children are overwritten. In a tree, however, overwriting becomes more harmful under our \textit{where} design above: since middle drafter nodes may be attached at earlier positions that already carry entire subtrees, overwriting would discard whole subtrees, including branches that the target model may still accept. We therefore need a non-destructive grafting rule, in which middle drafter nodes are attached beneath existing parents without removing any existing branches (Figure~\ref{fig:motivation}(c)).

\textbf{When to graft.} A middle drafter call pays off only at steps where its quality gain justifies the extra latency. Whether a step meets this condition mainly depends on the current tree, which is only known during decoding. We therefore need an online scheduler: at each step, it can read a few signals from the current tree and decide whether a middle drafter call is worth its cost; if so, the step is handed to the middle drafter, otherwise it stays with the small drafter (Figure~\ref{fig:motivation}(d)).

Building on this view, we propose \textit{TreeGraft}, a multi-drafter framework for tree-based speculative decoding. TreeGraft focuses on how multiple drafters jointly construct a shared draft tree and schedule its expansions. Our contributions are as follows: (1) \textbf{Expanded grafting-position selection}, which lets the middle drafter rescore previously under-ranked nodes and choose grafting positions; (2) \textbf{Non-destructive grafting}, which integrates middle drafter child nodes into the shared tree without discarding existing branches; (3) \textbf{Value-guided online scheduling}, which fits an offline value system to select value-predictive tree signals and distills its call/skip margin into a lightweight runtime scheduler; (4) Extensive experiments across 10 model pairs and 6 benchmarks, showing that TreeGraft outperforms the better of the two fixed single-drafter endpoint strategies by 15.1\% on average, reaching a maximum gain of 26.6\%.

%% file: sections/sec-2.tex
\section{Related Work}
\label{sec:related_work}
\textbf{Tree-based speculative decoding.}
Existing work on tree-based speculative decoding and its draft sources mainly improves candidate-tree quality from two perspectives.
The first is tree-side optimization, which improves tree construction~\citep{Li2024EAGLE2FI,Chen2024SequoiaSA}, node allocation or pruning~\citep{Wang2024OPTTreeSD,Zheng2025FasterSD}, and token or cache-tree verification~\citep{Miao2023SpecInferAL,Svirschevski2024SpecExecMP} under a given draft source.
\mbox{The second is draft-source improvement}, where representative training-based methods~\citep{https://doi.org/10.48550/arxiv.2505.12728,liu2026cats,https://doi.org/10.48550/arxiv.2601.18902,Cai2024MedusaSL,Ankner2024HydraSD,Li2024EAGLESS,Li2025EAGLE3} improve draft quality by training auxiliary draft modules or draft heads, including target-internal draft heads~\citep{Cai2024MedusaSL}, sequentially dependent draft heads~\citep{Ankner2024HydraSD}, feature-level self-drafting~\citep{Li2024EAGLESS}, and direct token prediction with multi-layer feature fusion~\citep{Li2025EAGLE3}.
TreeGraft extends the first direction from single-drafter tree construction to multi-drafter shared-tree construction.
At the same time, TreeGraft is largely orthogonal to the second direction, including training-based draft-source improvement, because it operates at the tree-construction level and does not constrain how each draft source is obtained.

\textbf{Hierarchical multi-drafter speculative decoding.}
Hierarchical multi-drafter speculative decoding augments the draft--target pipeline with a middle drafter, but existing methods are mainly sequence-based, such as Cascade Speculative Drafting~\citep{Chen2023CascadeSD} and TriForce~\citep{Sun2024TriForceLA}. In contrast, TreeGraft studies hierarchical multi-drafter speculative decoding on draft trees.

%% file: sections/sec-3.tex
\section{Shared-Tree Construction with Grafting}
\label{sec:grafting}

\subsection{Standard Single-Drafter Tree Expansion}
\label{subsec:single-drafter}

Before introducing grafting, we first describe the standard single-drafter expansion rule. At each drafting step, the drafter chooses a candidate pool $\mathcal{C}_k$ from the current draft tree and selects a frontier $\mathcal{F}_k\subseteq\mathcal{C}_k$ for expansion. In common single-drafter tree-SD methods~\citep{Li2024EAGLE2FI}, this candidate pool is restricted to the newly generated nodes from the previous step. We denote these nodes by $\mathcal{N}_k$, with $\mathcal{N}_1=\{r\}$ at the first drafting step, where $r$ is the root representing the current verified prefix. Thus, the standard single-drafter rule uses $\mathcal{C}_k=\mathcal{N}_k$. The drafter then ranks nodes in $\mathcal{C}_k$ using a cumulative path score $s(v)$ and selects the top-$W$ nodes as the expansion frontier, namely $\mathcal{F}_k=\mathrm{Top}_W(\mathcal{C}_k,s)$. Only nodes in $\mathcal{F}_k$ are expanded. For each frontier node, the drafter performs a forward pass on the prefix ending at that node, obtains the next-token distribution, and attaches the top-$m$ tokens as child nodes. These newly attached child nodes form $\mathcal{N}_{k+1}$ for the next step. By sharing prefixes among different paths, the resulting tree can be verified efficiently by the target model through tree attention.

The score $s(v)$ is defined as a cumulative probability along the path to node $v$. At drafting step $k$, let $\mathcal{T}_k$ denote the current tree and $V(\mathcal{T}_k)$ its node set. Each non-root node $u\in V(\mathcal{T}_k)\setminus\{r\}$ corresponds to a draft token generated under the prefix ending at its parent $\mathrm{pa}(u)$. When $u$ is generated, the drafter assigns a probability to this token; we store this probability as the edge score $p(u)$ of the parent-child edge leading to $u$. For a candidate node $v$, let $\pi(v)$ denote the nodes on the path from $r$ to $v$. The cumulative path score is $s(v)=\prod_{u\in\pi(v)\setminus\{r\}}p(u)$. For the root node, $s(r)=1$. 

\subsection{Grafting on a Shared Draft Tree}
\label{subsec:grafting}

The expansion process described in Section~\ref{subsec:single-drafter}
assumes a single drafter. When multiple drafters with varying
capabilities collaborate, they iteratively build upon a shared draft
tree. We frame this multi-drafter collaborative expansion as a
\textit{grafting} process. To formalize this operation, we must first
address two critical questions: \textit{where} to graft and \textit{how}
to graft, which together determine how TreeGraft improves tree quality.

\begin{figure}[!t]
    \centering
    \includegraphics[width=\linewidth]{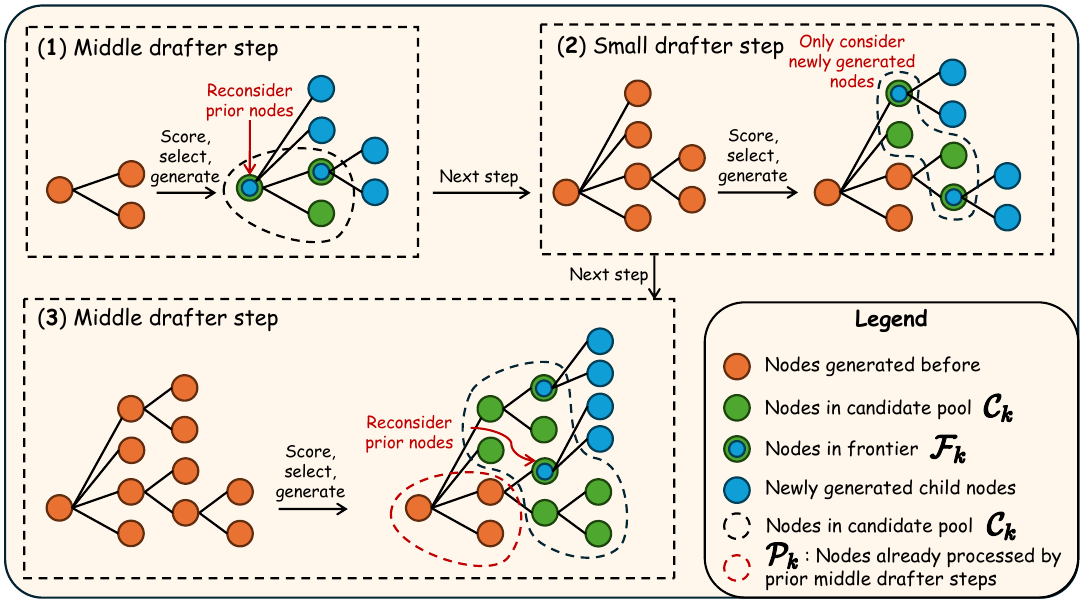}
    \caption{
    Illustration of expanded grafting-position selection under a
    middle--small--middle schedule. Panels \textbf{(1)} and \textbf{(3)}
    show that middle drafter steps reconsider earlier nodes in the shared
    tree, while panel \textbf{(2)} shows that the small drafter step only
    expands nodes generated in the previous step.
    }
    \label{fig:expanded_grafting_position}
\end{figure}

\textbf{Expanded grafting-position selection.}
To decide where to graft, TreeGraft looks at which nodes have been considered by the active drafter. We say that a drafter has \textit{visited} a node if the node has entered that drafter's candidate pool in a previous step. This definition recovers the standard rule in the
single-drafter case: since the same drafter has processed all earlier
candidate pools, the only nodes \textit{unvisited} by that drafter are the nodes
generated in the preceding step.

In TreeGraft, the two drafters are treated asymmetrically. The middle drafter follows the same \textit{visited}-node rule. Let $\mathcal{P}_k$ denote the set of nodes that
have already entered a middle drafter candidate pool before step $k$, with $\mathcal{P}_1 = \emptyset$.
At a middle drafter step, TreeGraft uses
$\mathcal{C}_k=V(\mathcal{T}_k)\setminus\mathcal{P}_k$. Thus, the middle drafter can
reconsider any node in the current shared tree that it has not yet
\textit{visited}, including historical nodes produced or ranked by the small drafter. Since the small drafter has lower capacity than the middle drafter, TreeGraft does not use it to \textit{revisit} older nodes for re-ranking. Instead, the small drafter treats all previous candidate pools, including those from the middle drafter, as already evaluated. Therefore, at a small drafter step, we set its candidate pool exclusively to the newly generated nodes from the previous step, i.e., $\mathcal{C}_k=\mathcal{N}_k$. After
a middle drafter step, we update
$\mathcal{P}_{k+1}=\mathcal{P}_k\cup\mathcal{C}_k$; after a small drafter
step, $\mathcal{P}_{k+1}=\mathcal{P}_k$. Combining the two cases gives
\begin{equation}
\mathcal{C}_k =
\begin{cases}
\mathcal{N}_k, & \text{small drafter step},\\
V(\mathcal{T}_k)\setminus\mathcal{P}_k, & \text{middle drafter step}.
\end{cases}
\end{equation}
Figure~\ref{fig:expanded_grafting_position} illustrates this rule under
a middle--small--middle schedule. Each middle drafter step considers
the portion of the shared tree that has not yet been \textit{visited} by the
middle drafter, while the small drafter expands only the most recently generated nodes. As a result, the middle drafter can select historical
grafting positions and correct earlier tree-growth decisions.

After the candidate pool is formed, we enforce the same asymmetry on scoring: the small drafter trusts and directly reuses the stored scores, whereas the middle drafter updates candidate scores using its own distribution. For each non-root candidate node $v\in\mathcal{C}_k$, the middle drafter updates the edge score $p(v)$ assigned to the token at $v$, and then recomputes the cumulative path score $s(v)$. The selection of the expansion frontier and the generation of child nodes then follow the same procedure as in Section~\ref{subsec:single-drafter}.

\textbf{Non-destructive grafting.}
Once the frontier $\mathcal{F}_k$ is selected, the drafter generates child nodes for each frontier node. A replacement rule is natural in sequence-based hierarchical drafting \citep{Chen2023CascadeSD, Sun2024TriForceLA}, where the stronger drafter refines a single continuation. In a shared tree, however, a selected historical node may already have children and descendants. Replacing them would remove existing subtrees before target verification and may reduce the accepted length (see
Appendix~\ref{app:nd-grafting-examples}). TreeGraft therefore attaches newly generated children under the frontier nodes without overwriting any existing node or branch.

Non-destructive grafting may temporarily make the constructed tree larger, but the final verification tree is pruned to the same verification budget \(B_{\mathrm{ver}}\). Once construction is complete, we prune the whole constructed tree before target verification. The pruning step ranks non-root nodes by cumulative path score. It then greedily keeps high-scoring nodes together with the ancestor chains needed to connect them to the root, until the verification budget is reached. Thus, old branches are not discarded during construction, while the final tree remains a valid budgeted tree for target-side verification.

\textbf{Distribution preservation.}
TreeGraft changes only the proposal tree supplied to the target model.
The final output is still produced by the same target-side tree
verification and rejection-sampling rule as in common single-drafter
tree-SD methods. Therefore, TreeGraft improves proposal-construction
efficiency without changing the target-model output distribution. Appendix~\ref{app:tree-construction-algorithms} gives the detailed pseudocode for standard single-drafter tree construction and TreeGraft shared-tree construction.

%% file: sections/sec-4.tex
\section{Online Scheduling via Value-Guided Distillation}
\label{sec:online_scheduling}

Section~\ref{sec:grafting} addresses where and how to graft middle drafter nodes onto the shared draft tree. We now address when to call the middle drafter. The value of a middle drafter call is state-dependent: in some tree states, the middle drafter can correct poor growth and increase the accepted length; in others, the same call adds latency without enough quality gain. TreeGraft therefore learns an online scheduler that maps the current online state to a call/skip decision. To obtain this scheduler, we first define the scheduling target, then fit an offline value system that connects trajectory-level outcomes to step-wise decisions, and finally distill its call/skip comparison into a lightweight runtime predictor, as summarized in Figure~\ref{fig:motivation}(d).

\subsection{Scheduling Target and Observability Gap}

A middle drafter call is useful only if the accepted-length gain it brings is worth the extra latency it adds. This makes throughput a natural objective for scheduling. The difficulty is that, when the scheduler makes a step-wise decision, this throughput is not yet available. It is observed only after the remaining tree construction and target-side verification are completed. Thus, the supervision for a step is tied to an entire scheduling trajectory rather than to an isolated local decision. To obtain step-wise supervision despite this observability gap, we fit an offline value system over complete trajectories and then distill its scheduling preference into a lightweight online scheduler.

Formally, turn-level throughput is written as $\eta=A/t$, where $A$ denotes the accepted length and $t=t_{\mathrm{D}}+t_{\mathrm{V}}$ is the total latency of one speculative-decoding turn, consisting of draft-construction latency $t_{\mathrm{D}}$ and target-verification latency $t_{\mathrm{V}}$. At drafting step $k$, the scheduler observes an online state $z_k$ and chooses a call/skip action $a_k\in\{0,1\}$: $a_k=1$ calls the middle drafter, while $a_k=0$ leaves the step to the small drafter. The desired scheduler chooses the action with the larger predicted turn-level throughput.

\subsection{Fitting an Offline Value System}

Since the final throughput depends on the entire sequence of scheduling decisions, we define a scheduling trajectory as the binary call/skip sequence $\tau_{1:D}\in\{0,1\}^D$, where $D$ is the maximum number of drafting steps and $\tau_k$ denotes the action taken at step $k$. For a step $k$, this trajectory can be decomposed into the past scheduling prefix $\tau_{1:k-1}$, the current action $a_k=\tau_k$, and the future suffix $\tau_{k+1:D}$. During offline data generation, we evaluate complete call/skip trajectories by executing the TreeGraft construction and target verification; Appendix~\ref{app:fitting_data} gives the data-collection details. Rather than compressing these evaluated outcomes into a single ratio $\eta=A/t$, we independently record the trajectory-level accepted length $A(\tau_{1:D})$ and latency $t(\tau_{1:D})$ as separate labels to provide more fine-grained supervision. To obtain step-wise supervision from these trajectory-level labels, we associate each complete-trajectory outcome with the pre-action state, current action, and future suffix observed at a step. Each fitting sample has the form $(z_k, a_k, \tau_{k+1:D}, A, t)$, where $A$ and $t$ are the complete-trajectory outcome labels paired with that step. We implement the value system $f_\theta$ as a lightweight MLP that predicts accepted length and latency from the current online state, the current action, and the future suffix, with architecture and training details in Appendix~\ref{app:value_system_fitting}:
\begin{equation}
(\hat{A}, \hat{t}) = f_\theta(z_k,a_k,\tau_{k+1:D}), \qquad
\hat{\eta} = \hat{A}/\hat{t}.
\end{equation}
Including the suffix is important: the same current state and current action can lead to different turn-level outcomes under different future schedules. Thus, $f_\theta$ serves as a future-aware estimator of the trajectory-level outcome induced by a local scheduling decision and its subsequent suffix.

The online state is written as $z_k = (\mathbf{c},k,\tau_{1:k-1},\phi_k)$, where $\mathbf{c}$ is the runtime context, $k$ is the drafting-step index, $\tau_{1:k-1}$ is the previous call/skip history, and $\phi_k$ contains tree signals computed from the current draft tree before taking action. The first three terms provide basic runtime and scheduling information, but they are insufficient to distinguish different tree states. Therefore, we additionally incorporate the tree signals $\phi_k$, which expose the current tree state to the value system.

Since the offline fitting samples are constructed from a limited set of models and datasets, directly using all candidate tree signals can introduce extra noise and reduce fitting accuracy. Not all tree signals $\phi_k$ are equally useful: some improve predictive accuracy, while others are redundant or noisy. To prevent this, we collect ten tree signals into a candidate set $\Phi_k$ and enumerate candidate subsets $\phi_k \subseteq \Phi_k$; Appendix~\ref{app:online_observations} defines the candidate signals. We combine each subset with the basic signals to formulate $z_k$, and fit the value system $f_\theta$. We select the subset $\phi_k^\star$ with the lowest out-of-fold identification score (detailed in Appendix~\ref{app:value_system_fitting}) as the final tree signals. This gives the final online state $z_k^\star$ and the corresponding fitted value system $f_{\theta^\star}$. With the selected state representation and fitted value system in place, the remaining question is how to convert this offline estimator into a runtime scheduling rule.

\subsection{Margin Distillation for Runtime Scheduling}

One possible scheduler would query the fitted value system online for both actions and choose the larger predicted throughput; Appendix~\ref{app:value-system-ablation} ablates this direct value-system planning alternative. We avoid this direct use for two reasons. First, the value system predicts the absolute accepted length and latency, while scheduling only needs the relative preference between call and skip. Small absolute prediction errors can therefore flip the action ordering. Second, the value system conditions on future suffixes, but future decisions are not known during online decoding.

TreeGraft instead uses the fitted value system as an offline teacher that converts trajectory-level outcomes into a call/skip margin. For each selected online state $z_k^\star$, we compare call and skip after optimizing over all binary future suffixes $\tau_{k+1:D}\in\{0,1\}^{D-k}$, as detailed in Appendix~\ref{app:runtime_policy}:
\begin{equation}
y_k =
\max_{\tau_{k+1:D}} \hat{\eta}(z_k^\star,1,\tau_{k+1:D})
-
\max_{\tau_{k+1:D}} \hat{\eta}(z_k^\star,0,\tau_{k+1:D}).
\end{equation}
The scalar $y_k$ is a call/skip margin. A positive margin means that calling the middle drafter is predicted to yield higher future-optimized throughput; a negative margin means that skipping is preferred. Since this target encodes the relative preference between call and skip at the same online
state, it can provide a more stable supervision signal than directly regressing complete-trajectory
throughput labels. We train a lightweight predictor $g_\psi(z_k^\star)$ to regress this margin. Unlike the value system, $g_\psi$ does not take the future suffix as input; the suffix-dependent reasoning has already been absorbed into the offline margin target. During inference, the scheduler uses only the sign of the predicted margin:
\begin{equation}
\pi_\psi(z_k^\star)=
\begin{cases}
1 \;(\text{call}), & g_\psi(z_k^\star)>0,\\
0 \;(\text{skip}), & \text{otherwise}.
\end{cases}
\end{equation}
Thus, online scheduling requires a single lightweight forward pass at each drafting step, while the expensive future-aware comparison is performed offline. Offline results of scheduler distillation are provided in Appendix~\ref{app:evaluation_results}. The runtime overhead of the scheduler is analyzed in Appendix~\ref{app:scheduler_runtime}.

%% file: sections/sec-5.tex
\section{Experiments}
\label{sec:experiments}

\subsection{Experimental Setup}
\label{sec:experimental_setup}

\textbf{Model pairs.} We evaluate TreeGraft with off-the-shelf drafters in a setting that requires \textit{no additional drafter training}.
The target model and the middle drafter are pretrained checkpoints from LLaMA~3~\citep{grattafiori2024llama3} and Qwen3~\citep{qwen3}.
\textit{For the small drafter, we use a training-free n-gram drafter.}
This choice is based on our preliminary experiments (Appendix~\ref{app:ngram_small_drafter}): even the smallest neural LLM in each model family introduces nontrivial forwarding cost in our implementation, making it unsuitable as the low-cost small drafter.
Among the preliminary configurations we tested, using a training-free n-gram drafter as the small drafter and a pretrained neural model as the middle drafter achieves the best speedup, while replacing the n-gram small drafter with a pretrained neural small model leads to lower speedup (Appendix~\ref{app:small drafter-choice}). The offline stage described in Section~\ref{sec:online_scheduling} (state selection, value-system fitting, and scheduler training) uses 6 target--middle model pairs.
In the online evaluation of the scheduler, we report results not only on these 6 seen pairs (white rows in Table~\ref{tab:main_results}), but also on 4 held-out pairs that are unseen during the offline stage (blue rows in Table~\ref{tab:main_results}), to verify the generalization.

\textbf{Datasets.} We evaluate on 6 benchmarks: GSM8K \citep{cobbe2021training}, Alpaca \citep{taori2023alpaca}, NQ \citep{kwiatkowski2019natural}, HumanEval \citep{chen2021evaluating}, CNN/DailyMail (CNN/DM) \citep{hermann2015teaching}, and MT-Bench \citep{zheng2023judging}. For the first five benchmarks, the offline stage and online evaluation use disjoint subsets, so no evaluation sample is used during offline fitting. MT-Bench, shown as the green columns in Table~\ref{tab:main_results}, is entirely held out from the offline stage and used only for cross-task generalization.

\textbf{Baselines and metrics.}
Under the standard dynamic-tree construction paradigm in tree-based speculative decoding~\citep{Li2024EAGLE2FI}, we construct two single-drafter endpoint baselines: All Small uses the small drafter throughout tree construction, while All Mid uses the middle drafter throughout. Comparing TreeGraft with these baselines tests whether selectively using multiple drafters to expand a shared tree improves over common single-drafter dynamic-tree construction. For fair comparison, all methods use the same tree budget, decoding parameters, and target verification procedure. In Section~\ref{sec:alternative_design_ablations}, we further compare with Cascade Speculative Drafting (CSD)~\citep{Chen2023CascadeSD}, a sequence-based multi-drafter baseline, and the standard dynamic-tree candidate-pool variant~\citep{Li2024EAGLE2FI} adapted to the multi-drafter setting.
We report the speedup ratio relative to standard autoregressive decoding with the target model.
Section~\ref{sec:alternative_design_ablations} additionally reports the average accepted length \(A\).

\textbf{Runtime configuration.} All online evaluations share the same tree-construction hyperparameters: maximum number of drafting steps $D{=}5$, expansion frontier size $W{=}10$, and $m{=}10$ child tokens per expanded node.
Additional decoding parameters are provided in Appendix~\ref{app:online_eval_setting_details}.

\subsection{Main Results}
\label{sec:main_results}

\textbf{Overall performance.}
Table~\ref{tab:main_results} compares TreeGraft with All Small and
All Mid, the two single-drafter dynamic-tree endpoints.
Across 10 model pairs and 6 datasets, TreeGraft achieves an average
speedup of \(1.60\times\), compared with \(1.32\times\) for All Small
and \(1.39\times\) for All Mid.
This represents a 15.1\% improvement in average speedup over All Mid,
the stronger of the two fixed endpoint strategies by overall average
speedup.
The largest per-setting gain over the stronger of the two endpoints is 26.6\%, attained on
Qwen3-32B/0.6B with Alpaca (\(1.38\times\) vs.\ \(1.09\times\)). Compared with the stronger endpoint selected separately for each
setting, TreeGraft performs better in 46 of the 60 model-pair--benchmark
settings, ties in two, and performs worse in 12.
In these 12 settings, its mean relative shortfall is 1.8\%.
Together, these results support the benefit of TreeGraft's
multi-drafter shared-tree construction over the two single-drafter
dynamic-tree endpoints in this controlled comparison.
Figure~\ref{fig:endpoint-comparison-heatmaps} visualizes the gains
over each endpoint, and Table~\ref{tab:full_main_results} in
Appendix~\ref{app:full-main-results} reports
the complete results.
Timing stability is examined in
Appendix~\ref{app:online-evaluation-stability}.

\newcommand{\fullmainresultstable}{%
\begin{table}[H]
\centering
\scriptsize
\setlength{\tabcolsep}{2.1pt}
\caption{Online evaluation results. Values are speedups over autoregressive target decoding. 
TG denotes TreeGraft, while S and M denote the single-drafter endpoints All Small and All Mid, 
respectively. The small drafter used by TG and S is a training-free n-gram drafter. 
Blue rows denote held-out target--middle pairs not used in offline fitting or scheduler 
distillation; green columns denote the held-out MT-Bench task; orange cells denote their 
intersection. For LLaMA pairs, 70B, 8B, 3B, and 1B refer to LLaMA 3.3-70B, LLaMA 3.1-8B, 
LLaMA 3.2-3B, and LLaMA 3.2-1B, respectively.}
\resizebox{\textwidth}{!}{
\begin{tabular}{l|ccc|ccc|ccc|ccc|ccc|ccc|ccc|cc}
\toprule
\multirow{2}{*}{Target--Middle}
& \multicolumn{3}{c|}{Alpaca}
& \multicolumn{3}{c|}{GSM8K}
& \multicolumn{3}{c|}{HumanEval}
& \multicolumn{3}{c|}{NQ}
& \multicolumn{3}{c|}{CNN/DM}
& \multicolumn{3}{c|}{\cellcolor{mtgreen}MT-Bench}
& \multicolumn{3}{c|}{Average}
& \multicolumn{2}{c}{Improvement} \\
\cmidrule(lr){2-4}
\cmidrule(lr){5-7}
\cmidrule(lr){8-10}
\cmidrule(lr){11-13}
\cmidrule(lr){14-16}
\cmidrule(lr){17-19}
\cmidrule(lr){20-22}
\cmidrule(lr){23-24}
& TG & S & M
& TG & S & M
& TG & S & M
& TG & S & M
& TG & S & M
& \cellcolor{mtgreen}TG & \cellcolor{mtgreen}S & \cellcolor{mtgreen}M
& TG & S & M
& TG vs S & TG vs M \\
\midrule
Qwen3-32B/1.7B
& \textbf{1.43} & 1.07 & 1.23
& \textbf{1.72} & 1.46 & 1.47
& \textbf{1.49} & 1.19 & 1.27
& \textbf{1.39} & 1.06 & 1.15
& \textbf{1.42} & 1.22 & 1.23
& \cellcolor{mtgreen}\textbf{1.51} & \cellcolor{mtgreen}1.26 & \cellcolor{mtgreen}1.21
& \textbf{1.49} & 1.21 & 1.26
& 23.1\% & 18.3\% \\

Qwen3-32B/0.6B
& \textbf{1.38} & 1.07 & 1.09
& \textbf{1.79} & 1.46 & 1.48
& \textbf{1.46} & 1.19 & 1.24
& \textbf{1.34} & 1.06 & 1.07
& \textbf{1.39} & 1.22 & 1.09
& \cellcolor{mtgreen}\textbf{1.46} & \cellcolor{mtgreen}1.26 & \cellcolor{mtgreen}1.18
& \textbf{1.47} & 1.21 & 1.19
& 21.5\% & 23.5\% \\

Qwen3-32B/8B
& \textbf{1.32} & 1.07 & 0.99
& \textbf{1.60} & 1.46 & 1.24
& \textbf{1.35} & 1.19 & 1.03
& \textbf{1.33} & 1.06 & 0.98
& \textbf{1.34} & 1.22 & 1.04
& \cellcolor{mtgreen}\textbf{1.43} & \cellcolor{mtgreen}1.26 & \cellcolor{mtgreen}1.08
& \textbf{1.40} & 1.21 & 1.06
& 15.7\% & 32.1\% \\

LLaMA-70B/1B
& \textbf{2.24} & 1.19 & 2.21
& 2.48 & 1.57 & \textbf{2.52}
& 2.52 & 1.82 & \textbf{2.58}
& \textbf{2.19} & 1.16 & 2.14
& 2.09 & 1.46 & \textbf{2.14}
& \cellcolor{mtgreen}\textbf{2.26} & \cellcolor{mtgreen}1.44 & \cellcolor{mtgreen}2.21
& \textbf{2.30} & 1.44 & \textbf{2.30}
& 59.7\% & 0.0\% \\

LLaMA-70B/3B
& \textbf{1.87} & 1.19 & \textbf{1.87}
& 2.05 & 1.57 & \textbf{2.08}
& 2.10 & 1.82 & \textbf{2.13}
& \textbf{1.89} & 1.16 & 1.84
& 1.85 & 1.46 & \textbf{1.87}
& \cellcolor{mtgreen}\textbf{1.91} & \cellcolor{mtgreen}1.44 & \cellcolor{mtgreen}\textbf{1.91}
& \textbf{1.95} & 1.44 & \textbf{1.95}
& 35.4\% & 0.0\% \\

LLaMA-8B/1B
& \textbf{1.28} & 1.14 & 1.27
& \textbf{1.59} & 1.50 & 1.37
& \textbf{1.87} & 1.75 & 1.45
& \textbf{1.33} & 1.15 & 1.21
& \textbf{1.48} & 1.44 & 1.25
& \cellcolor{mtgreen}\textbf{1.47} & \cellcolor{mtgreen}1.40 & \cellcolor{mtgreen}1.26
& \textbf{1.50} & 1.40 & 1.30
& 7.1\% & 15.4\% \\

\midrule
\rowcolor{holdoutblue}
Qwen3-32B/4B
& \textbf{1.33} & 1.07 & 1.10
& \textbf{1.61} & 1.46 & 1.30
& \textbf{1.31} & 1.19 & 1.15
& \textbf{1.31} & 1.06 & 1.08
& \textbf{1.34} & 1.22 & 1.13
& \cellcolor{intersectorange}\textbf{1.41} & \cellcolor{intersectorange}1.26 & \cellcolor{intersectorange}1.15
& \textbf{1.39} & 1.21 & 1.15
& 14.9\% & 20.9\% \\

\rowcolor{holdoutblue}
Qwen3-8B/0.6B
& \textbf{1.14} & 1.10 & 0.80
& \textbf{1.50} & 1.45 & 0.97
& \textbf{1.24} & 1.23 & 0.89
& \textbf{1.19} & 1.16 & 0.80
& \textbf{1.24} & 1.19 & 0.80
& \cellcolor{intersectorange}\textbf{1.29} & \cellcolor{intersectorange}1.28 & \cellcolor{intersectorange}0.84
& \textbf{1.27} & 1.24 & 0.85
& 2.4\% & 49.4\% \\

\rowcolor{holdoutblue}
LLaMA-70B/8B
& 1.77 & 1.19 & \textbf{1.78}
& 1.93 & 1.57 & \textbf{1.97}
& 1.93 & 1.82 & \textbf{2.03}
& 1.73 & 1.16 & \textbf{1.75}
& 1.77 & 1.46 & \textbf{1.82}
& \cellcolor{intersectorange}\textbf{1.83} & \cellcolor{intersectorange}1.44 & \cellcolor{intersectorange}1.81
& 1.83 & 1.44 & \textbf{1.86}
& 27.1\% & -1.6\% \\

\rowcolor{holdoutblue}
LLaMA-8B/3B
& \textbf{1.17} & 1.14 & 0.91
& \textbf{1.53} & 1.50 & 0.97
& 1.74 & \textbf{1.75} & 1.02
& \textbf{1.19} & 1.15 & 0.89
& \textbf{1.46} & 1.44 & 0.92
& \cellcolor{intersectorange}\textbf{1.42} & \cellcolor{intersectorange}1.40 & \cellcolor{intersectorange}0.94
& \textbf{1.42} & 1.40 & 0.94
& 1.4\% & 51.1\% \\

\midrule
Average
& \textbf{1.49} & 1.12 & 1.33
& \textbf{1.78} & 1.50 & 1.54
& \textbf{1.70} & 1.50 & 1.48
& \textbf{1.49} & 1.12 & 1.29
& \textbf{1.54} & 1.33 & 1.33
& \cellcolor{mtgreen}\textbf{1.60} & \cellcolor{mtgreen}1.34 & \cellcolor{mtgreen}1.36
& \textbf{1.60} & 1.32 & 1.39
& 21.2\% & 15.1\% \\
\bottomrule
\end{tabular}
}
\label{tab:full_main_results}
\end{table}
}

\begin{table*}[!t]
\centering
\small
\setlength{\tabcolsep}{4pt}
\caption{Online evaluation results. Values are speedups over autoregressive target decoding. TG denotes TreeGraft; S and M denote All Small and All Mid. The five-task average is the arithmetic mean over Alpaca, GSM8K, HumanEval, NQ, and CNN/DM; the overall average includes MT-Bench. Blue rows denote held-out target--middle pairs, green columns denote the held-out MT-Bench task, and orange cells denote their intersection. For LLaMA pairs, 70B, 8B, 3B, and 1B refer to LLaMA 3.3-70B, LLaMA 3.1-8B, LLaMA 3.2-3B, and LLaMA 3.2-1B, respectively. The full speedups are in Table~\ref{tab:full_main_results}.}
\label{tab:main_results}
\resizebox{\textwidth}{!}{%
\begin{tabular}{l|ccc|ccc|ccc|cc}
\toprule
\multirow{2}{*}{Target--Middle}
& \multicolumn{3}{c|}{Five-task average}
& \multicolumn{3}{c|}{\cellcolor{mtgreen}MT-Bench}
& \multicolumn{3}{c|}{Overall average}
& \multicolumn{2}{c}{Improvement} \\
\cmidrule(lr){2-4}
\cmidrule(lr){5-7}
\cmidrule(lr){8-10}
\cmidrule(lr){11-12}
& TG & S & M
& \cellcolor{mtgreen}TG & \cellcolor{mtgreen}S & \cellcolor{mtgreen}M
& TG & S & M
& TG vs S & TG vs M \\
\midrule
Qwen3-32B/1.7B
& \textbf{1.49} & 1.20 & 1.27
& \cellcolor{mtgreen}\textbf{1.51} & \cellcolor{mtgreen}1.26 & \cellcolor{mtgreen}1.21
& \textbf{1.49} & 1.21 & 1.26
& 23.1\% & 18.3\% \\
Qwen3-32B/0.6B
& \textbf{1.47} & 1.20 & 1.19
& \cellcolor{mtgreen}\textbf{1.46} & \cellcolor{mtgreen}1.26 & \cellcolor{mtgreen}1.18
& \textbf{1.47} & 1.21 & 1.19
& 21.5\% & 23.5\% \\
Qwen3-32B/8B
& \textbf{1.39} & 1.20 & 1.06
& \cellcolor{mtgreen}\textbf{1.43} & \cellcolor{mtgreen}1.26 & \cellcolor{mtgreen}1.08
& \textbf{1.40} & 1.21 & 1.06
& 15.7\% & 32.1\% \\
LLaMA-70B/1B
& 2.30 & 1.44 & \textbf{2.32}
& \cellcolor{mtgreen}\textbf{2.26} & \cellcolor{mtgreen}1.44 & \cellcolor{mtgreen}2.21
& \textbf{2.30} & 1.44 & \textbf{2.30}
& 59.7\% & 0.0\% \\
LLaMA-70B/3B
& 1.95 & 1.44 & \textbf{1.96}
& \cellcolor{mtgreen}\textbf{1.91} & \cellcolor{mtgreen}1.44 & \cellcolor{mtgreen}\textbf{1.91}
& \textbf{1.95} & 1.44 & \textbf{1.95}
& 35.4\% & 0.0\% \\
LLaMA-8B/1B
& \textbf{1.51} & 1.40 & 1.31
& \cellcolor{mtgreen}\textbf{1.47} & \cellcolor{mtgreen}1.40 & \cellcolor{mtgreen}1.26
& \textbf{1.50} & 1.40 & 1.30
& 7.1\% & 15.4\% \\
\midrule
\rowcolor{holdoutblue}
Qwen3-32B/4B
& \textbf{1.38} & 1.20 & 1.15
& \cellcolor{intersectorange}\textbf{1.41} & \cellcolor{intersectorange}1.26 & \cellcolor{intersectorange}1.15
& \textbf{1.39} & 1.21 & 1.15
& 14.9\% & 20.9\% \\
\rowcolor{holdoutblue}
Qwen3-8B/0.6B
& \textbf{1.26} & 1.23 & 0.85
& \cellcolor{intersectorange}\textbf{1.29} & \cellcolor{intersectorange}1.28 & \cellcolor{intersectorange}0.84
& \textbf{1.27} & 1.24 & 0.85
& 2.4\% & 49.4\% \\
\rowcolor{holdoutblue}
LLaMA-70B/8B
& 1.83 & 1.44 & \textbf{1.87}
& \cellcolor{intersectorange}\textbf{1.83} & \cellcolor{intersectorange}1.44 & \cellcolor{intersectorange}1.81
& 1.83 & 1.44 & \textbf{1.86}
& 27.1\% & -1.6\% \\
\rowcolor{holdoutblue}
LLaMA-8B/3B
& \textbf{1.42} & 1.40 & 0.94
& \cellcolor{intersectorange}\textbf{1.42} & \cellcolor{intersectorange}1.40 & \cellcolor{intersectorange}0.94
& \textbf{1.42} & 1.40 & 0.94
& 1.4\% & 51.1\% \\
\midrule
Average
& \textbf{1.60} & 1.31 & 1.39
& \cellcolor{mtgreen}\textbf{1.60} & \cellcolor{mtgreen}1.34 & \cellcolor{mtgreen}1.36
& \textbf{1.60} & 1.32 & 1.39
& 21.2\% & 15.1\% \\
\bottomrule
\end{tabular}
}
\end{table*}

\begin{figure*}[!t]
\centering
\input{figures/endpoint_comparison_heatmaps.tex}
\vspace{-14pt}
\caption{Setting-wise percentage speedup gains of TreeGraft over All Small (left) and All Mid (right). Cell labels show gains computed from the reported two-decimal speedups. An asterisk (*) marks held-out model pairs and MT-Bench.}
\label{fig:endpoint-comparison-heatmaps}
\end{figure*}

\begin{minipage}{\textwidth}
\textbf{Generalization.}
TreeGraft maintains speedup on unseen model pairs and tasks.
The 4 held-out model pairs (blue rows in Table~\ref{tab:main_results}) achieve an average speedup of $1.48\times$, higher than All Small ($1.32\times$) and All Mid ($1.20\times$).
On MT-Bench, shown as the green columns in Table~\ref{tab:main_results} and excluded from the offline stage, TreeGraft achieves an average speedup of $1.60\times$, higher than All~Small ($1.34\times$) and All~Mid ($1.36\times$).
These results suggest that the scheduler captures cost--quality trade-off patterns that transfer
across model pairs and tasks, rather than being tied only to the fitting configurations.
We also evaluate scheduler transfer without refitting across sampling temperature ($T=0\rightarrow1$), tree width, model family, and hardware (Appendix~\ref{app:sampling-transfer}).
\end{minipage}

\begin{figure*}[!t]
\centering
\includegraphics[width=\textwidth]{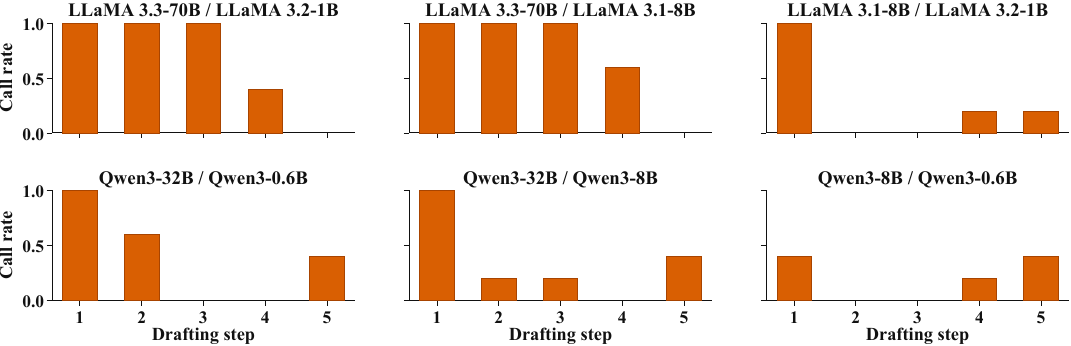}
\caption{Step-wise middle drafter usage learned by the TreeGraft online scheduler under the value system setting.
Each subplot corresponds to one target--middle model pair.
The y-axis shows the middle drafter call rate at each drafting step, averaged over the five fit datasets.
Higher values indicate a stronger tendency to call the middle drafter at that step.}
\label{fig:call_distribution}
\end{figure*}

\textbf{Scheduling behavior analysis.}
At the model-pair-average level, TreeGraft improves over the better endpoint by up to 21.5\% (Qwen3-32B/0.6B, TreeGraft $1.47\times$ vs.\ All~Small $1.21\times$), while falling behind the better endpoint by at most 1.6\% (LLaMA 3.3-70B/3.1-8B, TreeGraft $1.83\times$ vs.\ All~Mid $1.86\times$).
The former indicates that the scheduler effectively captures gains when improvement is available; the latter indicates that the scheduler introduces almost no loss when improvement is unavailable.
Together, these results indicate that the scheduler adaptively finds a favorable cost--quality trade-off across settings, thereby achieving the goal of building high-quality trees at low cost.
The following analysis explains this adaptive behavior through three representative regimes:

When All Mid significantly outperforms All Small (e.g., LLaMA 3.3-70B/3.2-1B, All Mid 2.30$\times$
vs. All Small 1.44$\times$), TreeGraft achieves 2.30$\times$ speedup, closely tracking the All Mid endpoint.
As shown in Figure~\ref{fig:call_distribution}, the learned scheduler calls the middle drafter frequently
for most drafting steps in this regime, yielding a call pattern close to All Mid.

When All Small significantly outperforms All Mid (e.g., Qwen3-8B/0.6B, All Small 1.24$\times$ vs.
All Mid 0.85$\times$), TreeGraft achieves 1.27$\times$ speedup, slightly exceeding All Small while avoiding
the severe slowdown of All Mid. In this case, Figure~\ref{fig:call_distribution} shows that the scheduler
uses the middle drafter sparsely, keeping the call rate much lower than an All Mid policy.

When the two endpoints achieve similar and modest speedups (e.g., Qwen3-32B/0.6B, All Small
1.21$\times$ vs. All Mid 1.19$\times$), TreeGraft achieves 1.47$\times$ speedup, substantially surpassing both
endpoints. Here, Figure~\ref{fig:call_distribution} shows a selective call pattern: middle drafter calls
are concentrated on selected steps rather than applied uniformly.

Overall, these call-rate patterns are consistent with the intended adaptive cost--quality trade-off:
the scheduler adjusts middle-drafter usage across model pairs and drafting steps, increasing calls
when stronger drafting is more beneficial and reducing them when its cost is less justified.

\subsection{Alternative Designs and Ablations}
\label{sec:alternative_design_ablations}

\begin{table}[H]
\centering 
\caption{Ablation of grafting mechanism components. All variants use the same fixed middle drafter schedule $[1,0,1,0,1]$, i.e., the middle drafter participates only at drafting steps 1, 3, and 5. Speedup is relative to autoregressive decoding with the target model. $A$ denotes average accepted length.}
\label{tab:ablation}
\scriptsize
\setlength{\tabcolsep}{3pt}
\resizebox{\textwidth}{!}{%
\begin{tabular}{l|cc|cc|cc|cc|cc}
\toprule
\multirow{2}{*}{Method}
& \multicolumn{2}{c|}{Q32B/0.6B GSM8K}
& \multicolumn{2}{c|}{Q32B/0.6B CNN/DM}
& \multicolumn{2}{c|}{L70B/1B GSM8K}
& \multicolumn{2}{c|}{L70B/1B CNN/DM}
& \multicolumn{2}{c}{Average} \\
\cmidrule(lr){2-3} \cmidrule(lr){4-5} \cmidrule(lr){6-7} \cmidrule(lr){8-9} \cmidrule(lr){10-11}
& Speedup ($\times$) & $A$ & Speedup ($\times$) & $A$ & Speedup ($\times$) & $A$ & Speedup ($\times$) & $A$ & Speedup ($\times$) & $A$ \\
\midrule
CSD       & 0.75 & 0.82 & 0.78 & 0.72 & 1.07 & 0.90 & 1.14 & 0.78 & 0.94 & 0.81 \\
w/o EG    & 1.49 & 1.76 & 0.85 & 1.56 & 1.72 & 1.92 & 1.34 & 1.97 & 1.35 & 1.80 \\
w/o ND    & 1.44 & 1.69 & 0.82 & 1.18 & 1.33 & 1.64 & 1.21 & 1.37 & 1.20 & 1.47 \\
TreeGraft & \textbf{1.55} & \textbf{1.84} & \textbf{0.95} & \textbf{2.15} & \textbf{1.80} & \textbf{2.18} & \textbf{1.42} & \textbf{2.52} & \textbf{1.43} & \textbf{2.17} \\
\bottomrule
\end{tabular}
}
\end{table}

\begin{minipage}{\textwidth}
This section ablates TreeGraft's three components: expanded grafting-position selection and non-destructive grafting (how middle drafter outputs are integrated) and the online scheduler (which steps invoke the middle drafter). We ablate the first two components under a fixed schedule, and then assess the scheduler by comparing the fixed-schedule results with the learned-scheduler results in Table~\ref{tab:main_results}.
\end{minipage}

\textbf{Setup.}
All variants share a fixed middle drafter schedule $[1,0,1,0,1]$, where the middle drafter participates at steps 1, 3, and 5, so grafting is tested after small drafter expansion at multiple tree depths. We use two model pairs (Qwen3-32B/0.6B, LLaMA 3.3-70B/3.2-1B), each selected from a representative regime where middle drafter calls are effective (Section~\ref{sec:main_results}). Experiments are conducted on GSM8K and CNN/DM. The four methods are: \textbf{ Cascade Speculative Drafting (CSD)} \citep{Chen2023CascadeSD}, a sequence-based multi-drafter baseline reproduced under the same fixed middle drafter schedule; \textbf{w/o EG} (w/o expanded grafting-position selection), which retains non-destructive grafting but restricts the middle drafter candidate pool to the nodes generated in the previous step (this structure effectively implements the EAGLE candidate-pool design within a multi-drafter setting); \textbf{w/o ND} (w/o non-destructive grafting), which retains expanded grafting-position selection but overwrites existing children when attaching middle drafter nodes; and the full \textbf{TreeGraft}.

Table~\ref{tab:ablation} reports the speedup and average accepted length (tokens accepted per verification round). TreeGraft achieves the best results across all settings.

\textbf{Tree-based grafting outperforms CSD.}
Under the same fixed middle drafter schedule, replacing TreeGraft with CSD reduces the overall speedup from $1.43\times$ to $0.94\times$ and the average accepted length from 2.17 to 0.81, showing that shared-tree grafting is more effective than sequence-based multi-drafter methods when call positions are controlled.

\textbf{Expanded grafting-position selection provides an additional, independent gain.}
Removing it (w/o EG) reduces the overall speedup from $1.43\times$ to $1.35\times$ and the average accepted length from 2.17 to 1.80. To check whether this gain indeed comes from revisiting earlier nodes, we report detailed statistics at middle drafter steps: TreeGraft's candidate pool averages 112.54 nodes ($2.38\times$ that of w/o EG), and $61.47\%$ of its frontier selections fall on historical positions (vs.\ $0\%$ for w/o EG). Since both methods select the same top-$W$ nodes per step, the accepted-length improvement stems from better positions, not more selections.

\textbf{Non-destructive grafting provides the primary gain.}
Removing it (w/o ND) reduces the overall speedup from $1.43\times$ to $1.20\times$ and the average accepted length from 2.17 to 1.47, showing that overwriting existing branches can remove useful paths and thus reduce target acceptance.

\textbf{Contribution of the online scheduler.}
Comparing the fixed schedule with the same settings in Table~\ref{tab:main_results},
the online scheduler raises the average speedup from $1.43\times$ to $1.94\times$.
The gap is sharpest on CNN/DM, where it improves Qwen from $0.95\times$ to $1.39\times$ and LLaMA from $1.42\times$ to $2.09\times$, confirming that adaptive scheduling provides a gain beyond the grafting mechanism itself.

\textbf{Scheduler-state ablation.} Removing runtime context, scheduling history, or tree signals separately reduces performance, showing that call cost, earlier drafting choices, and current frontier structure each inform the decision. In particular, tree signals contribute information beyond timing and call history (Appendix~\ref{app:scheduler-feature-ablation}).

\textbf{Heuristic scheduler comparison.} We test simple policies based on step position, confidence, score gap, or latency, tuning each over a predefined parameter grid. The learned scheduler attains the highest average speedup among these tested policies (Appendix~\ref{app:heuristic-schedulers}).

%% file: figures/endpoint_comparison_heatmaps.tex
\begin{tikzpicture}[x=1in,y=1in]
\path[use as bounding box] (0,0.06) rectangle (5.5,3.56);
\node[font=\fontsize{7}{7.7}\selectfont\bfseries] at (1.940,3.440) {TG vs. All Small};
\node[anchor=west,rotate=45,inner sep=0pt,font=\fontsize{6.5}{7.1}\selectfont\bfseries] at (0.827,2.970) {Alpaca};
\node[anchor=west,rotate=45,inner sep=0pt,font=\fontsize{6.5}{7.1}\selectfont\bfseries] at (1.200,2.970) {GSM8K};
\node[anchor=west,rotate=45,inner sep=0pt,font=\fontsize{6.5}{7.1}\selectfont\bfseries] at (1.573,2.970) {HumanEval};
\node[anchor=west,rotate=45,inner sep=0pt,font=\fontsize{6.5}{7.1}\selectfont\bfseries] at (1.947,2.970) {NQ};
\node[anchor=west,rotate=45,inner sep=0pt,font=\fontsize{6.5}{7.1}\selectfont\bfseries] at (2.320,2.970) {CNN/DM};
\node[anchor=west,rotate=45,inner sep=0pt,font=\fontsize{6.5}{7.1}\selectfont\bfseries] at (2.693,2.970) {MT-Bench*};
\node[anchor=east,font=\fontsize{6.5}{7.1}\selectfont\bfseries] at (0.775,2.778) {Qwen3-32B/1.7B};
\definecolor{hmCell82r0c0}{HTML}{E7914A}
\fill[hmCell82r0c0] (0.824,2.670) rectangle (1.189,2.886);
\node[text=black,font=\fontsize{6.5}{7.1}\selectfont\bfseries] at (1.007,2.778) {+33.6};
\definecolor{hmCell82r0c1}{HTML}{F0C194}
\fill[hmCell82r0c1] (1.197,2.670) rectangle (1.563,2.886);
\node[text=black,font=\fontsize{6.5}{7.1}\selectfont\bfseries] at (1.380,2.778) {+17.8};
\definecolor{hmCell82r0c2}{HTML}{ECA86C}
\fill[hmCell82r0c2] (1.571,2.670) rectangle (1.936,2.886);
\node[text=black,font=\fontsize{6.5}{7.1}\selectfont\bfseries] at (1.753,2.778) {+25.2};
\definecolor{hmCell82r0c3}{HTML}{E99854}
\fill[hmCell82r0c3] (1.944,2.670) rectangle (2.309,2.886);
\node[text=black,font=\fontsize{6.5}{7.1}\selectfont\bfseries] at (2.127,2.778) {+31.1};
\definecolor{hmCell82r0c4}{HTML}{F0C59B}
\fill[hmCell82r0c4] (2.317,2.670) rectangle (2.683,2.886);
\node[text=black,font=\fontsize{6.5}{7.1}\selectfont\bfseries] at (2.500,2.778) {+16.4};
\definecolor{hmCell82r0c5}{HTML}{EFBA89}
\fill[hmCell82r0c5] (2.691,2.670) rectangle (3.056,2.886);
\node[text=black,font=\fontsize{6.5}{7.1}\selectfont\bfseries] at (2.873,2.778) {+19.8};
\node[anchor=east,font=\fontsize{6.5}{7.1}\selectfont\bfseries] at (0.775,2.554) {Qwen3-32B/0.6B};
\definecolor{hmCell82r1c0}{HTML}{EA9E5D}
\fill[hmCell82r1c0] (0.824,2.446) rectangle (1.189,2.662);
\node[text=black,font=\fontsize{6.5}{7.1}\selectfont\bfseries] at (1.007,2.554) {+29.0};
\definecolor{hmCell82r1c1}{HTML}{EDB17A}
\fill[hmCell82r1c1] (1.197,2.446) rectangle (1.563,2.662);
\node[text=black,font=\fontsize{6.5}{7.1}\selectfont\bfseries] at (1.380,2.554) {+22.6};
\definecolor{hmCell82r1c2}{HTML}{EDB179}
\fill[hmCell82r1c2] (1.571,2.446) rectangle (1.936,2.662);
\node[text=black,font=\fontsize{6.5}{7.1}\selectfont\bfseries] at (1.753,2.554) {+22.7};
\definecolor{hmCell82r1c3}{HTML}{EBA567}
\fill[hmCell82r1c3] (1.944,2.446) rectangle (2.309,2.662);
\node[text=black,font=\fontsize{6.5}{7.1}\selectfont\bfseries] at (2.127,2.554) {+26.4};
\definecolor{hmCell82r1c4}{HTML}{F2CEAA}
\fill[hmCell82r1c4] (2.317,2.446) rectangle (2.683,2.662);
\node[text=black,font=\fontsize{6.5}{7.1}\selectfont\bfseries] at (2.500,2.554) {+13.9};
\definecolor{hmCell82r1c5}{HTML}{F1C79E}
\fill[hmCell82r1c5] (2.691,2.446) rectangle (3.056,2.662);
\node[text=black,font=\fontsize{6.5}{7.1}\selectfont\bfseries] at (2.873,2.554) {+15.9};
\node[anchor=east,font=\fontsize{6.5}{7.1}\selectfont\bfseries] at (0.775,2.330) {Qwen3-32B/8B};
\definecolor{hmCell82r2c0}{HTML}{EDAE76}
\fill[hmCell82r2c0] (0.824,2.222) rectangle (1.189,2.438);
\node[text=black,font=\fontsize{6.5}{7.1}\selectfont\bfseries] at (1.007,2.330) {+23.4};
\definecolor{hmCell82r2c1}{HTML}{F5DEC4}
\fill[hmCell82r2c1] (1.197,2.222) rectangle (1.563,2.438);
\node[text=black,font=\fontsize{6.5}{7.1}\selectfont\bfseries] at (1.380,2.330) {+9.6};
\definecolor{hmCell82r2c2}{HTML}{F2D0AD}
\fill[hmCell82r2c2] (1.571,2.222) rectangle (1.936,2.438);
\node[text=black,font=\fontsize{6.5}{7.1}\selectfont\bfseries] at (1.753,2.330) {+13.4};
\definecolor{hmCell82r2c3}{HTML}{ECA86B}
\fill[hmCell82r2c3] (1.944,2.222) rectangle (2.309,2.438);
\node[text=black,font=\fontsize{6.5}{7.1}\selectfont\bfseries] at (2.127,2.330) {+25.5};
\definecolor{hmCell82r2c4}{HTML}{F5DDC3}
\fill[hmCell82r2c4] (2.317,2.222) rectangle (2.683,2.438);
\node[text=black,font=\fontsize{6.5}{7.1}\selectfont\bfseries] at (2.500,2.330) {+9.8};
\definecolor{hmCell82r2c5}{HTML}{F2D0AC}
\fill[hmCell82r2c5] (2.691,2.222) rectangle (3.056,2.438);
\node[text=black,font=\fontsize{6.5}{7.1}\selectfont\bfseries] at (2.873,2.330) {+13.5};
\node[anchor=east,font=\fontsize{6.5}{7.1}\selectfont\bfseries] at (0.775,2.106) {LLaMA-70B/1B};
\definecolor{hmCell82r3c0}{HTML}{D96003}
\fill[hmCell82r3c0] (0.824,1.998) rectangle (1.189,2.214);
\node[text=black,font=\fontsize{6.5}{7.1}\selectfont\bfseries] at (1.007,2.106) {+88.2};
\definecolor{hmCell82r3c1}{HTML}{DF6F18}
\fill[hmCell82r3c1] (1.197,1.998) rectangle (1.563,2.214);
\node[text=black,font=\fontsize{6.5}{7.1}\selectfont\bfseries] at (1.380,2.106) {+58.0};
\definecolor{hmCell82r3c2}{HTML}{E58337}
\fill[hmCell82r3c2] (1.571,1.998) rectangle (1.936,2.214);
\node[text=black,font=\fontsize{6.5}{7.1}\selectfont\bfseries] at (1.753,2.106) {+38.5};
\definecolor{hmCell82r3c3}{HTML}{D96003}
\fill[hmCell82r3c3] (1.944,1.998) rectangle (2.309,2.214);
\node[text=black,font=\fontsize{6.5}{7.1}\selectfont\bfseries] at (2.127,2.106) {+88.8};
\definecolor{hmCell82r3c4}{HTML}{E37C2D}
\fill[hmCell82r3c4] (2.317,1.998) rectangle (2.683,2.214);
\node[text=black,font=\fontsize{6.5}{7.1}\selectfont\bfseries] at (2.500,2.106) {+43.2};
\definecolor{hmCell82r3c5}{HTML}{DF7019}
\fill[hmCell82r3c5] (2.691,1.998) rectangle (3.056,2.214);
\node[text=black,font=\fontsize{6.5}{7.1}\selectfont\bfseries] at (2.873,2.106) {+56.9};
\node[anchor=east,font=\fontsize{6.5}{7.1}\selectfont\bfseries] at (0.775,1.882) {LLaMA-70B/3B};
\definecolor{hmCell82r4c0}{HTML}{DF7019}
\fill[hmCell82r4c0] (0.824,1.774) rectangle (1.189,1.990);
\node[text=black,font=\fontsize{6.5}{7.1}\selectfont\bfseries] at (1.007,1.882) {+57.1};
\definecolor{hmCell82r4c1}{HTML}{E99957}
\fill[hmCell82r4c1] (1.197,1.774) rectangle (1.563,1.990);
\node[text=black,font=\fontsize{6.5}{7.1}\selectfont\bfseries] at (1.380,1.882) {+30.6};
\definecolor{hmCell82r4c2}{HTML}{F1C9A1}
\fill[hmCell82r4c2] (1.571,1.774) rectangle (1.936,1.990);
\node[text=black,font=\fontsize{6.5}{7.1}\selectfont\bfseries] at (1.753,1.882) {+15.4};
\definecolor{hmCell82r4c3}{HTML}{DE6C13}
\fill[hmCell82r4c3] (1.944,1.774) rectangle (2.309,1.990);
\node[text=black,font=\fontsize{6.5}{7.1}\selectfont\bfseries] at (2.127,1.882) {+62.9};
\definecolor{hmCell82r4c4}{HTML}{EBA466}
\fill[hmCell82r4c4] (2.317,1.774) rectangle (2.683,1.990);
\node[text=black,font=\fontsize{6.5}{7.1}\selectfont\bfseries] at (2.500,1.882) {+26.7};
\definecolor{hmCell82r4c5}{HTML}{E8944E}
\fill[hmCell82r4c5] (2.691,1.774) rectangle (3.056,1.990);
\node[text=black,font=\fontsize{6.5}{7.1}\selectfont\bfseries] at (2.873,1.882) {+32.6};
\node[anchor=east,font=\fontsize{6.5}{7.1}\selectfont\bfseries] at (0.775,1.658) {LLaMA-8B/1B};
\definecolor{hmCell82r5c0}{HTML}{F3D4B4}
\fill[hmCell82r5c0] (0.824,1.550) rectangle (1.189,1.766);
\node[text=black,font=\fontsize{6.5}{7.1}\selectfont\bfseries] at (1.007,1.658) {+12.3};
\definecolor{hmCell82r5c1}{HTML}{F7E9D6}
\fill[hmCell82r5c1] (1.197,1.550) rectangle (1.563,1.766);
\node[text=black,font=\fontsize{6.5}{7.1}\selectfont\bfseries] at (1.380,1.658) {+6.0};
\definecolor{hmCell82r5c2}{HTML}{F7E6D2}
\fill[hmCell82r5c2] (1.571,1.550) rectangle (1.936,1.766);
\node[text=black,font=\fontsize{6.5}{7.1}\selectfont\bfseries] at (1.753,1.658) {+6.9};
\definecolor{hmCell82r5c3}{HTML}{F1C89F}
\fill[hmCell82r5c3] (1.944,1.550) rectangle (2.309,1.766);
\node[text=black,font=\fontsize{6.5}{7.1}\selectfont\bfseries] at (2.127,1.658) {+15.7};
\definecolor{hmCell82r5c4}{HTML}{F8F0E5}
\fill[hmCell82r5c4] (2.317,1.550) rectangle (2.683,1.766);
\node[text=black,font=\fontsize{6.5}{7.1}\selectfont\bfseries] at (2.500,1.658) {+2.8};
\definecolor{hmCell82r5c5}{HTML}{F8ECDB}
\fill[hmCell82r5c5] (2.691,1.550) rectangle (3.056,1.766);
\node[text=black,font=\fontsize{6.5}{7.1}\selectfont\bfseries] at (2.873,1.658) {+5.0};
\node[anchor=east,font=\fontsize{6.5}{7.1}\selectfont\bfseries] at (0.775,1.434) {Qwen3-32B/4B*};
\definecolor{hmCell82r6c0}{HTML}{ECAB71}
\fill[hmCell82r6c0] (0.824,1.326) rectangle (1.189,1.542);
\node[text=black,font=\fontsize{6.5}{7.1}\selectfont\bfseries] at (1.007,1.434) {+24.3};
\definecolor{hmCell82r6c1}{HTML}{F5DCC0}
\fill[hmCell82r6c1] (1.197,1.326) rectangle (1.563,1.542);
\node[text=black,font=\fontsize{6.5}{7.1}\selectfont\bfseries] at (1.380,1.434) {+10.3};
\definecolor{hmCell82r6c2}{HTML}{F5DDC1}
\fill[hmCell82r6c2] (1.571,1.326) rectangle (1.936,1.542);
\node[text=black,font=\fontsize{6.5}{7.1}\selectfont\bfseries] at (1.753,1.434) {+10.1};
\definecolor{hmCell82r6c3}{HTML}{EDAE75}
\fill[hmCell82r6c3] (1.944,1.326) rectangle (2.309,1.542);
\node[text=black,font=\fontsize{6.5}{7.1}\selectfont\bfseries] at (2.127,1.434) {+23.6};
\definecolor{hmCell82r6c4}{HTML}{F5DDC3}
\fill[hmCell82r6c4] (2.317,1.326) rectangle (2.683,1.542);
\node[text=black,font=\fontsize{6.5}{7.1}\selectfont\bfseries] at (2.500,1.434) {+9.8};
\definecolor{hmCell82r6c5}{HTML}{F3D6B6}
\fill[hmCell82r6c5] (2.691,1.326) rectangle (3.056,1.542);
\node[text=black,font=\fontsize{6.5}{7.1}\selectfont\bfseries] at (2.873,1.434) {+11.9};
\node[anchor=east,font=\fontsize{6.5}{7.1}\selectfont\bfseries] at (0.775,1.210) {Qwen3-8B/0.6B*};
\definecolor{hmCell82r7c0}{HTML}{F8EFE1}
\fill[hmCell82r7c0] (0.824,1.102) rectangle (1.189,1.318);
\node[text=black,font=\fontsize{6.5}{7.1}\selectfont\bfseries] at (1.007,1.210) {+3.6};
\definecolor{hmCell82r7c1}{HTML}{F8EFE2}
\fill[hmCell82r7c1] (1.197,1.102) rectangle (1.563,1.318);
\node[text=black,font=\fontsize{6.5}{7.1}\selectfont\bfseries] at (1.380,1.210) {+3.4};
\definecolor{hmCell82r7c2}{HTML}{F8F4EE}
\fill[hmCell82r7c2] (1.571,1.102) rectangle (1.936,1.318);
\node[text=black,font=\fontsize{6.5}{7.1}\selectfont\bfseries] at (1.753,1.210) {+0.8};
\definecolor{hmCell82r7c3}{HTML}{F8F1E6}
\fill[hmCell82r7c3] (1.944,1.102) rectangle (2.309,1.318);
\node[text=black,font=\fontsize{6.5}{7.1}\selectfont\bfseries] at (2.127,1.210) {+2.6};
\definecolor{hmCell82r7c4}{HTML}{F8EEDF}
\fill[hmCell82r7c4] (2.317,1.102) rectangle (2.683,1.318);
\node[text=black,font=\fontsize{6.5}{7.1}\selectfont\bfseries] at (2.500,1.210) {+4.2};
\definecolor{hmCell82r7c5}{HTML}{F8F4EE}
\fill[hmCell82r7c5] (2.691,1.102) rectangle (3.056,1.318);
\node[text=black,font=\fontsize{6.5}{7.1}\selectfont\bfseries] at (2.873,1.210) {+0.8};
\node[anchor=east,font=\fontsize{6.5}{7.1}\selectfont\bfseries] at (0.775,0.986) {LLaMA-70B/8B*};
\definecolor{hmCell82r8c0}{HTML}{E17725}
\fill[hmCell82r8c0] (0.824,0.878) rectangle (1.189,1.094);
\node[text=black,font=\fontsize{6.5}{7.1}\selectfont\bfseries] at (1.007,0.986) {+48.7};
\definecolor{hmCell82r8c1}{HTML}{EDB078}
\fill[hmCell82r8c1] (1.197,0.878) rectangle (1.563,1.094);
\node[text=black,font=\fontsize{6.5}{7.1}\selectfont\bfseries] at (1.380,0.986) {+22.9};
\definecolor{hmCell82r8c2}{HTML}{F7E9D6}
\fill[hmCell82r8c2] (1.571,0.878) rectangle (1.936,1.094);
\node[text=black,font=\fontsize{6.5}{7.1}\selectfont\bfseries] at (1.753,0.986) {+6.0};
\definecolor{hmCell82r8c3}{HTML}{E17724}
\fill[hmCell82r8c3] (1.944,0.878) rectangle (2.309,1.094);
\node[text=black,font=\fontsize{6.5}{7.1}\selectfont\bfseries] at (2.127,0.986) {+49.1};
\definecolor{hmCell82r8c4}{HTML}{EEB581}
\fill[hmCell82r8c4] (2.317,0.878) rectangle (2.683,1.094);
\node[text=black,font=\fontsize{6.5}{7.1}\selectfont\bfseries] at (2.500,0.986) {+21.2};
\definecolor{hmCell82r8c5}{HTML}{EBA365}
\fill[hmCell82r8c5] (2.691,0.878) rectangle (3.056,1.094);
\node[text=black,font=\fontsize{6.5}{7.1}\selectfont\bfseries] at (2.873,0.986) {+27.1};
\node[anchor=east,font=\fontsize{6.5}{7.1}\selectfont\bfseries] at (0.775,0.762) {LLaMA-8B/3B*};
\definecolor{hmCell82r9c0}{HTML}{F8F1E6}
\fill[hmCell82r9c0] (0.824,0.654) rectangle (1.189,0.870);
\node[text=black,font=\fontsize{6.5}{7.1}\selectfont\bfseries] at (1.007,0.762) {+2.6};
\definecolor{hmCell82r9c1}{HTML}{F8F2E9}
\fill[hmCell82r9c1] (1.197,0.654) rectangle (1.563,0.870);
\node[text=black,font=\fontsize{6.5}{7.1}\selectfont\bfseries] at (1.380,0.762) {+2.0};
\definecolor{hmCell82r9c2}{HTML}{F7F5F1}
\fill[hmCell82r9c2] (1.571,0.654) rectangle (1.936,0.870);
\node[text=black,font=\fontsize{6.5}{7.1}\selectfont\bfseries] at (1.753,0.762) {-0.6};
\definecolor{hmCell82r9c3}{HTML}{F8EFE2}
\fill[hmCell82r9c3] (1.944,0.654) rectangle (2.309,0.870);
\node[text=black,font=\fontsize{6.5}{7.1}\selectfont\bfseries] at (2.127,0.762) {+3.5};
\definecolor{hmCell82r9c4}{HTML}{F8F3EC}
\fill[hmCell82r9c4] (2.317,0.654) rectangle (2.683,0.870);
\node[text=black,font=\fontsize{6.5}{7.1}\selectfont\bfseries] at (2.500,0.762) {+1.4};
\definecolor{hmCell82r9c5}{HTML}{F8F3EB}
\fill[hmCell82r9c5] (2.691,0.654) rectangle (3.056,0.870);
\node[text=black,font=\fontsize{6.5}{7.1}\selectfont\bfseries] at (2.873,0.762) {+1.4};
\node[font=\fontsize{7}{7.7}\selectfont\bfseries] at (4.320,3.440) {TG vs. All Mid};
\node[anchor=west,rotate=45,inner sep=0pt,font=\fontsize{6.5}{7.1}\selectfont\bfseries] at (3.207,2.970) {Alpaca};
\node[anchor=west,rotate=45,inner sep=0pt,font=\fontsize{6.5}{7.1}\selectfont\bfseries] at (3.580,2.970) {GSM8K};
\node[anchor=west,rotate=45,inner sep=0pt,font=\fontsize{6.5}{7.1}\selectfont\bfseries] at (3.953,2.970) {HumanEval};
\node[anchor=west,rotate=45,inner sep=0pt,font=\fontsize{6.5}{7.1}\selectfont\bfseries] at (4.327,2.970) {NQ};
\node[anchor=west,rotate=45,inner sep=0pt,font=\fontsize{6.5}{7.1}\selectfont\bfseries] at (4.700,2.970) {CNN/DM};
\node[anchor=west,rotate=45,inner sep=0pt,font=\fontsize{6.5}{7.1}\selectfont\bfseries] at (5.073,2.970) {MT-Bench*};
\definecolor{hmCell320r0c0}{HTML}{F0C69C}
\fill[hmCell320r0c0] (3.204,2.670) rectangle (3.569,2.886);
\node[text=black,font=\fontsize{6.5}{7.1}\selectfont\bfseries] at (3.387,2.778) {+16.3};
\definecolor{hmCell320r0c1}{HTML}{F0C398}
\fill[hmCell320r0c1] (3.577,2.670) rectangle (3.943,2.886);
\node[text=black,font=\fontsize{6.5}{7.1}\selectfont\bfseries] at (3.760,2.778) {+17.0};
\definecolor{hmCell320r0c2}{HTML}{F0C296}
\fill[hmCell320r0c2] (3.951,2.670) rectangle (4.316,2.886);
\node[text=black,font=\fontsize{6.5}{7.1}\selectfont\bfseries] at (4.133,2.778) {+17.3};
\definecolor{hmCell320r0c3}{HTML}{EEB783}
\fill[hmCell320r0c3] (4.324,2.670) rectangle (4.689,2.886);
\node[text=black,font=\fontsize{6.5}{7.1}\selectfont\bfseries] at (4.507,2.778) {+20.9};
\definecolor{hmCell320r0c4}{HTML}{F1C9A1}
\fill[hmCell320r0c4] (4.697,2.670) rectangle (5.063,2.886);
\node[text=black,font=\fontsize{6.5}{7.1}\selectfont\bfseries] at (4.880,2.778) {+15.4};
\definecolor{hmCell320r0c5}{HTML}{ECAA6E}
\fill[hmCell320r0c5] (5.071,2.670) rectangle (5.436,2.886);
\node[text=black,font=\fontsize{6.5}{7.1}\selectfont\bfseries] at (5.253,2.778) {+24.8};
\definecolor{hmCell320r1c0}{HTML}{EBA567}
\fill[hmCell320r1c0] (3.204,2.446) rectangle (3.569,2.662);
\node[text=black,font=\fontsize{6.5}{7.1}\selectfont\bfseries] at (3.387,2.554) {+26.6};
\definecolor{hmCell320r1c1}{HTML}{EEB683}
\fill[hmCell320r1c1] (3.577,2.446) rectangle (3.943,2.662);
\node[text=black,font=\fontsize{6.5}{7.1}\selectfont\bfseries] at (3.760,2.554) {+20.9};
\definecolor{hmCell320r1c2}{HTML}{F0C194}
\fill[hmCell320r1c2] (3.951,2.446) rectangle (4.316,2.662);
\node[text=black,font=\fontsize{6.5}{7.1}\selectfont\bfseries] at (4.133,2.554) {+17.7};
\definecolor{hmCell320r1c3}{HTML}{ECA86C}
\fill[hmCell320r1c3] (4.324,2.446) rectangle (4.689,2.662);
\node[text=black,font=\fontsize{6.5}{7.1}\selectfont\bfseries] at (4.507,2.554) {+25.2};
\definecolor{hmCell320r1c4}{HTML}{EBA263}
\fill[hmCell320r1c4] (4.697,2.446) rectangle (5.063,2.662);
\node[text=black,font=\fontsize{6.5}{7.1}\selectfont\bfseries] at (4.880,2.554) {+27.5};
\definecolor{hmCell320r1c5}{HTML}{EDAD74}
\fill[hmCell320r1c5] (5.071,2.446) rectangle (5.436,2.662);
\node[text=black,font=\fontsize{6.5}{7.1}\selectfont\bfseries] at (5.253,2.554) {+23.7};
\definecolor{hmCell320r2c0}{HTML}{E8924C}
\fill[hmCell320r2c0] (3.204,2.222) rectangle (3.569,2.438);
\node[text=black,font=\fontsize{6.5}{7.1}\selectfont\bfseries] at (3.387,2.330) {+33.3};
\definecolor{hmCell320r2c1}{HTML}{EA9E5D}
\fill[hmCell320r2c1] (3.577,2.222) rectangle (3.943,2.438);
\node[text=black,font=\fontsize{6.5}{7.1}\selectfont\bfseries] at (3.760,2.330) {+29.0};
\definecolor{hmCell320r2c2}{HTML}{E99855}
\fill[hmCell320r2c2] (3.951,2.222) rectangle (4.316,2.438);
\node[text=black,font=\fontsize{6.5}{7.1}\selectfont\bfseries] at (4.133,2.330) {+31.1};
\definecolor{hmCell320r2c3}{HTML}{E68B42}
\fill[hmCell320r2c3] (4.324,2.222) rectangle (4.689,2.438);
\node[text=black,font=\fontsize{6.5}{7.1}\selectfont\bfseries] at (4.507,2.330) {+35.7};
\definecolor{hmCell320r2c4}{HTML}{EA9E5E}
\fill[hmCell320r2c4] (4.697,2.222) rectangle (5.063,2.438);
\node[text=black,font=\fontsize{6.5}{7.1}\selectfont\bfseries] at (4.880,2.330) {+28.8};
\definecolor{hmCell320r2c5}{HTML}{E8944F}
\fill[hmCell320r2c5] (5.071,2.222) rectangle (5.436,2.438);
\node[text=black,font=\fontsize{6.5}{7.1}\selectfont\bfseries] at (5.253,2.330) {+32.4};
\definecolor{hmCell320r3c0}{HTML}{F8F3EC}
\fill[hmCell320r3c0] (3.204,1.998) rectangle (3.569,2.214);
\node[text=black,font=\fontsize{6.5}{7.1}\selectfont\bfseries] at (3.387,2.106) {+1.4};
\definecolor{hmCell320r3c1}{HTML}{F5F4F0}
\fill[hmCell320r3c1] (3.577,1.998) rectangle (3.943,2.214);
\node[text=black,font=\fontsize{6.5}{7.1}\selectfont\bfseries] at (3.760,2.106) {-1.6};
\definecolor{hmCell320r3c2}{HTML}{F4F3EF}
\fill[hmCell320r3c2] (3.951,1.998) rectangle (4.316,2.214);
\node[text=black,font=\fontsize{6.5}{7.1}\selectfont\bfseries] at (4.133,2.106) {-2.3};
\definecolor{hmCell320r3c3}{HTML}{F8F1E7}
\fill[hmCell320r3c3] (4.324,1.998) rectangle (4.689,2.214);
\node[text=black,font=\fontsize{6.5}{7.1}\selectfont\bfseries] at (4.507,2.106) {+2.3};
\definecolor{hmCell320r3c4}{HTML}{F4F2EF}
\fill[hmCell320r3c4] (4.697,1.998) rectangle (5.063,2.214);
\node[text=black,font=\fontsize{6.5}{7.1}\selectfont\bfseries] at (4.880,2.106) {-2.3};
\definecolor{hmCell320r3c5}{HTML}{F8F1E8}
\fill[hmCell320r3c5] (5.071,1.998) rectangle (5.436,2.214);
\node[text=black,font=\fontsize{6.5}{7.1}\selectfont\bfseries] at (5.253,2.106) {+2.3};
\definecolor{hmCell320r4c0}{HTML}{F8F6F2}
\fill[hmCell320r4c0] (3.204,1.774) rectangle (3.569,1.990);
\node[text=black,font=\fontsize{6.5}{7.1}\selectfont\bfseries] at (3.387,1.882) {0.0};
\definecolor{hmCell320r4c1}{HTML}{F5F4F0}
\fill[hmCell320r4c1] (3.577,1.774) rectangle (3.943,1.990);
\node[text=black,font=\fontsize{6.5}{7.1}\selectfont\bfseries] at (3.760,1.882) {-1.4};
\definecolor{hmCell320r4c2}{HTML}{F5F4F0}
\fill[hmCell320r4c2] (3.951,1.774) rectangle (4.316,1.990);
\node[text=black,font=\fontsize{6.5}{7.1}\selectfont\bfseries] at (4.133,1.882) {-1.4};
\definecolor{hmCell320r4c3}{HTML}{F8F1E6}
\fill[hmCell320r4c3] (4.324,1.774) rectangle (4.689,1.990);
\node[text=black,font=\fontsize{6.5}{7.1}\selectfont\bfseries] at (4.507,1.882) {+2.7};
\definecolor{hmCell320r4c4}{HTML}{F6F4F1}
\fill[hmCell320r4c4] (4.697,1.774) rectangle (5.063,1.990);
\node[text=black,font=\fontsize{6.5}{7.1}\selectfont\bfseries] at (4.880,1.882) {-1.1};
\definecolor{hmCell320r4c5}{HTML}{F8F6F2}
\fill[hmCell320r4c5] (5.071,1.774) rectangle (5.436,1.990);
\node[text=black,font=\fontsize{6.5}{7.1}\selectfont\bfseries] at (5.253,1.882) {0.0};
\definecolor{hmCell320r5c0}{HTML}{F8F4EE}
\fill[hmCell320r5c0] (3.204,1.550) rectangle (3.569,1.766);
\node[text=black,font=\fontsize{6.5}{7.1}\selectfont\bfseries] at (3.387,1.658) {+0.8};
\definecolor{hmCell320r5c1}{HTML}{F0C79D}
\fill[hmCell320r5c1] (3.577,1.550) rectangle (3.943,1.766);
\node[text=black,font=\fontsize{6.5}{7.1}\selectfont\bfseries] at (3.760,1.658) {+16.1};
\definecolor{hmCell320r5c2}{HTML}{EA9E5D}
\fill[hmCell320r5c2] (3.951,1.550) rectangle (4.316,1.766);
\node[text=black,font=\fontsize{6.5}{7.1}\selectfont\bfseries] at (4.133,1.658) {+29.0};
\definecolor{hmCell320r5c3}{HTML}{F5DDC2}
\fill[hmCell320r5c3] (4.324,1.550) rectangle (4.689,1.766);
\node[text=black,font=\fontsize{6.5}{7.1}\selectfont\bfseries] at (4.507,1.658) {+9.9};
\definecolor{hmCell320r5c4}{HTML}{EFBF91}
\fill[hmCell320r5c4] (4.697,1.550) rectangle (5.063,1.766);
\node[text=black,font=\fontsize{6.5}{7.1}\selectfont\bfseries] at (4.880,1.658) {+18.4};
\definecolor{hmCell320r5c5}{HTML}{F0C49A}
\fill[hmCell320r5c5] (5.071,1.550) rectangle (5.436,1.766);
\node[text=black,font=\fontsize{6.5}{7.1}\selectfont\bfseries] at (5.253,1.658) {+16.7};
\definecolor{hmCell320r6c0}{HTML}{EEB783}
\fill[hmCell320r6c0] (3.204,1.326) rectangle (3.569,1.542);
\node[text=black,font=\fontsize{6.5}{7.1}\selectfont\bfseries] at (3.387,1.434) {+20.9};
\definecolor{hmCell320r6c1}{HTML}{EDAD73}
\fill[hmCell320r6c1] (3.577,1.326) rectangle (3.943,1.542);
\node[text=black,font=\fontsize{6.5}{7.1}\selectfont\bfseries] at (3.760,1.434) {+23.8};
\definecolor{hmCell320r6c2}{HTML}{F2CEAA}
\fill[hmCell320r6c2] (3.951,1.326) rectangle (4.316,1.542);
\node[text=black,font=\fontsize{6.5}{7.1}\selectfont\bfseries] at (4.133,1.434) {+13.9};
\definecolor{hmCell320r6c3}{HTML}{EEB581}
\fill[hmCell320r6c3] (4.324,1.326) rectangle (4.689,1.542);
\node[text=black,font=\fontsize{6.5}{7.1}\selectfont\bfseries] at (4.507,1.434) {+21.3};
\definecolor{hmCell320r6c4}{HTML}{EFBE90}
\fill[hmCell320r6c4] (4.697,1.326) rectangle (5.063,1.542);
\node[text=black,font=\fontsize{6.5}{7.1}\selectfont\bfseries] at (4.880,1.434) {+18.6};
\definecolor{hmCell320r6c5}{HTML}{EDB17A}
\fill[hmCell320r6c5] (5.071,1.326) rectangle (5.436,1.542);
\node[text=black,font=\fontsize{6.5}{7.1}\selectfont\bfseries] at (5.253,1.434) {+22.6};
\definecolor{hmCell320r7c0}{HTML}{E37D2E}
\fill[hmCell320r7c0] (3.204,1.102) rectangle (3.569,1.318);
\node[text=black,font=\fontsize{6.5}{7.1}\selectfont\bfseries] at (3.387,1.210) {+42.5};
\definecolor{hmCell320r7c1}{HTML}{E0721D}
\fill[hmCell320r7c1] (3.577,1.102) rectangle (3.943,1.318);
\node[text=black,font=\fontsize{6.5}{7.1}\selectfont\bfseries] at (3.760,1.210) {+54.6};
\definecolor{hmCell320r7c2}{HTML}{E48134}
\fill[hmCell320r7c2] (3.951,1.102) rectangle (4.316,1.318);
\node[text=black,font=\fontsize{6.5}{7.1}\selectfont\bfseries] at (4.133,1.210) {+39.3};
\definecolor{hmCell320r7c3}{HTML}{E17725}
\fill[hmCell320r7c3] (4.324,1.102) rectangle (4.689,1.318);
\node[text=black,font=\fontsize{6.5}{7.1}\selectfont\bfseries] at (4.507,1.210) {+48.7};
\definecolor{hmCell320r7c4}{HTML}{E0721C}
\fill[hmCell320r7c4] (4.697,1.102) rectangle (5.063,1.318);
\node[text=black,font=\fontsize{6.5}{7.1}\selectfont\bfseries] at (4.880,1.210) {+55.0};
\definecolor{hmCell320r7c5}{HTML}{E0731E}
\fill[hmCell320r7c5] (5.071,1.102) rectangle (5.436,1.318);
\node[text=black,font=\fontsize{6.5}{7.1}\selectfont\bfseries] at (5.253,1.210) {+53.6};
\definecolor{hmCell320r8c0}{HTML}{F7F5F1}
\fill[hmCell320r8c0] (3.204,0.878) rectangle (3.569,1.094);
\node[text=black,font=\fontsize{6.5}{7.1}\selectfont\bfseries] at (3.387,0.986) {-0.6};
\definecolor{hmCell320r8c1}{HTML}{F4F3EF}
\fill[hmCell320r8c1] (3.577,0.878) rectangle (3.943,1.094);
\node[text=black,font=\fontsize{6.5}{7.1}\selectfont\bfseries] at (3.760,0.986) {-2.0};
\definecolor{hmCell320r8c2}{HTML}{EFEFEB}
\fill[hmCell320r8c2] (3.951,0.878) rectangle (4.316,1.094);
\node[text=black,font=\fontsize{6.5}{7.1}\selectfont\bfseries] at (4.133,0.986) {-4.9};
\definecolor{hmCell320r8c3}{HTML}{F6F4F0}
\fill[hmCell320r8c3] (4.324,0.878) rectangle (4.689,1.094);
\node[text=black,font=\fontsize{6.5}{7.1}\selectfont\bfseries] at (4.507,0.986) {-1.1};
\definecolor{hmCell320r8c4}{HTML}{F3F2EE}
\fill[hmCell320r8c4] (4.697,0.878) rectangle (5.063,1.094);
\node[text=black,font=\fontsize{6.5}{7.1}\selectfont\bfseries] at (4.880,0.986) {-2.7};
\definecolor{hmCell320r8c5}{HTML}{F8F4ED}
\fill[hmCell320r8c5] (5.071,0.878) rectangle (5.436,1.094);
\node[text=black,font=\fontsize{6.5}{7.1}\selectfont\bfseries] at (5.253,0.986) {+1.1};
\definecolor{hmCell320r9c0}{HTML}{EA9F5F}
\fill[hmCell320r9c0] (3.204,0.654) rectangle (3.569,0.870);
\node[text=black,font=\fontsize{6.5}{7.1}\selectfont\bfseries] at (3.387,0.762) {+28.6};
\definecolor{hmCell320r9c1}{HTML}{DF6F18}
\fill[hmCell320r9c1] (3.577,0.654) rectangle (3.943,0.870);
\node[text=black,font=\fontsize{6.5}{7.1}\selectfont\bfseries] at (3.760,0.762) {+57.7};
\definecolor{hmCell320r9c2}{HTML}{DC680E}
\fill[hmCell320r9c2] (3.951,0.654) rectangle (4.316,0.870);
\node[text=black,font=\fontsize{6.5}{7.1}\selectfont\bfseries] at (4.133,0.762) {+70.6};
\definecolor{hmCell320r9c3}{HTML}{E7914A}
\fill[hmCell320r9c3] (4.324,0.654) rectangle (4.689,0.870);
\node[text=black,font=\fontsize{6.5}{7.1}\selectfont\bfseries] at (4.507,0.762) {+33.7};
\definecolor{hmCell320r9c4}{HTML}{DE6E17}
\fill[hmCell320r9c4] (4.697,0.654) rectangle (5.063,0.870);
\node[text=black,font=\fontsize{6.5}{7.1}\selectfont\bfseries] at (4.880,0.762) {+58.7};
\definecolor{hmCell320r9c5}{HTML}{E17522}
\fill[hmCell320r9c5] (5.071,0.654) rectangle (5.436,0.870);
\node[text=black,font=\fontsize{6.5}{7.1}\selectfont\bfseries] at (5.253,0.762) {+51.1};
\node[font=\fontsize{7}{7.7}\selectfont\bfseries] at (2.750,0.515) {Relative speedup gain (\%)};
\definecolor{hmLegend0}{HTML}{E6E7E4}
\fill[hmLegend0] (1.3000,0.290) rectangle (1.3186,0.390);
\definecolor{hmLegend1}{HTML}{E7E8E5}
\fill[hmLegend1] (1.3181,0.290) rectangle (1.3368,0.390);
\definecolor{hmLegend2}{HTML}{E8E9E6}
\fill[hmLegend2] (1.3363,0.290) rectangle (1.3549,0.390);
\definecolor{hmLegend3}{HTML}{E9EAE7}
\fill[hmLegend3] (1.3544,0.290) rectangle (1.3730,0.390);
\definecolor{hmLegend4}{HTML}{EAEBE8}
\fill[hmLegend4] (1.3725,0.290) rectangle (1.3911,0.390);
\definecolor{hmLegend5}{HTML}{ECECE9}
\fill[hmLegend5] (1.3906,0.290) rectangle (1.4092,0.390);
\definecolor{hmLegend6}{HTML}{EDEDEA}
\fill[hmLegend6] (1.4087,0.290) rectangle (1.4274,0.390);
\definecolor{hmLegend7}{HTML}{EEEEEB}
\fill[hmLegend7] (1.4269,0.290) rectangle (1.4455,0.390);
\definecolor{hmLegend8}{HTML}{EFEFEB}
\fill[hmLegend8] (1.4450,0.290) rectangle (1.4636,0.390);
\definecolor{hmLegend9}{HTML}{F0F0EC}
\fill[hmLegend9] (1.4631,0.290) rectangle (1.4818,0.390);
\definecolor{hmLegend10}{HTML}{F1F1ED}
\fill[hmLegend10] (1.4813,0.290) rectangle (1.4999,0.390);
\definecolor{hmLegend11}{HTML}{F3F2EE}
\fill[hmLegend11] (1.4994,0.290) rectangle (1.5180,0.390);
\definecolor{hmLegend12}{HTML}{F4F3EF}
\fill[hmLegend12] (1.5175,0.290) rectangle (1.5361,0.390);
\definecolor{hmLegend13}{HTML}{F5F4F0}
\fill[hmLegend13] (1.5356,0.290) rectangle (1.5542,0.390);
\definecolor{hmLegend14}{HTML}{F6F5F1}
\fill[hmLegend14] (1.5537,0.290) rectangle (1.5724,0.390);
\definecolor{hmLegend15}{HTML}{F7F6F2}
\fill[hmLegend15] (1.5719,0.290) rectangle (1.5905,0.390);
\definecolor{hmLegend16}{HTML}{F8F5F1}
\fill[hmLegend16] (1.5900,0.290) rectangle (1.6086,0.390);
\definecolor{hmLegend17}{HTML}{F8F4EE}
\fill[hmLegend17] (1.6081,0.290) rectangle (1.6268,0.390);
\definecolor{hmLegend18}{HTML}{F8F3EB}
\fill[hmLegend18] (1.6263,0.290) rectangle (1.6449,0.390);
\definecolor{hmLegend19}{HTML}{F8F2E8}
\fill[hmLegend19] (1.6444,0.290) rectangle (1.6630,0.390);
\definecolor{hmLegend20}{HTML}{F8F0E5}
\fill[hmLegend20] (1.6625,0.290) rectangle (1.6811,0.390);
\definecolor{hmLegend21}{HTML}{F8EFE2}
\fill[hmLegend21] (1.6806,0.290) rectangle (1.6992,0.390);
\definecolor{hmLegend22}{HTML}{F8EEDF}
\fill[hmLegend22] (1.6987,0.290) rectangle (1.7174,0.390);
\definecolor{hmLegend23}{HTML}{F8EDDC}
\fill[hmLegend23] (1.7169,0.290) rectangle (1.7355,0.390);
\definecolor{hmLegend24}{HTML}{F8EBD9}
\fill[hmLegend24] (1.7350,0.290) rectangle (1.7536,0.390);
\definecolor{hmLegend25}{HTML}{F7E9D6}
\fill[hmLegend25] (1.7531,0.290) rectangle (1.7718,0.390);
\definecolor{hmLegend26}{HTML}{F7E7D3}
\fill[hmLegend26] (1.7713,0.290) rectangle (1.7899,0.390);
\definecolor{hmLegend27}{HTML}{F7E5D0}
\fill[hmLegend27] (1.7894,0.290) rectangle (1.8080,0.390);
\definecolor{hmLegend28}{HTML}{F6E4CD}
\fill[hmLegend28] (1.8075,0.290) rectangle (1.8261,0.390);
\definecolor{hmLegend29}{HTML}{F6E2CA}
\fill[hmLegend29] (1.8256,0.290) rectangle (1.8442,0.390);
\definecolor{hmLegend30}{HTML}{F6E0C7}
\fill[hmLegend30] (1.8438,0.290) rectangle (1.8624,0.390);
\definecolor{hmLegend31}{HTML}{F5DEC4}
\fill[hmLegend31] (1.8619,0.290) rectangle (1.8805,0.390);
\definecolor{hmLegend32}{HTML}{F5DCC0}
\fill[hmLegend32] (1.8800,0.290) rectangle (1.8986,0.390);
\definecolor{hmLegend33}{HTML}{F4D9BC}
\fill[hmLegend33] (1.8981,0.290) rectangle (1.9168,0.390);
\definecolor{hmLegend34}{HTML}{F4D7B8}
\fill[hmLegend34] (1.9163,0.290) rectangle (1.9349,0.390);
\definecolor{hmLegend35}{HTML}{F3D5B4}
\fill[hmLegend35] (1.9344,0.290) rectangle (1.9530,0.390);
\definecolor{hmLegend36}{HTML}{F3D2B1}
\fill[hmLegend36] (1.9525,0.290) rectangle (1.9711,0.390);
\definecolor{hmLegend37}{HTML}{F2D0AD}
\fill[hmLegend37] (1.9706,0.290) rectangle (1.9892,0.390);
\definecolor{hmLegend38}{HTML}{F2CEA9}
\fill[hmLegend38] (1.9888,0.290) rectangle (2.0074,0.390);
\definecolor{hmLegend39}{HTML}{F1CBA5}
\fill[hmLegend39] (2.0069,0.290) rectangle (2.0255,0.390);
\definecolor{hmLegend40}{HTML}{F1C9A1}
\fill[hmLegend40] (2.0250,0.290) rectangle (2.0436,0.390);
\definecolor{hmLegend41}{HTML}{F1C79E}
\fill[hmLegend41] (2.0431,0.290) rectangle (2.0618,0.390);
\definecolor{hmLegend42}{HTML}{F0C59B}
\fill[hmLegend42] (2.0613,0.290) rectangle (2.0799,0.390);
\definecolor{hmLegend43}{HTML}{F0C397}
\fill[hmLegend43] (2.0794,0.290) rectangle (2.0980,0.390);
\definecolor{hmLegend44}{HTML}{F0C194}
\fill[hmLegend44] (2.0975,0.290) rectangle (2.1161,0.390);
\definecolor{hmLegend45}{HTML}{EFBF90}
\fill[hmLegend45] (2.1156,0.290) rectangle (2.1343,0.390);
\definecolor{hmLegend46}{HTML}{EFBD8D}
\fill[hmLegend46] (2.1338,0.290) rectangle (2.1524,0.390);
\definecolor{hmLegend47}{HTML}{EFBB8A}
\fill[hmLegend47] (2.1519,0.290) rectangle (2.1705,0.390);
\definecolor{hmLegend48}{HTML}{EEB886}
\fill[hmLegend48] (2.1700,0.290) rectangle (2.1886,0.390);
\definecolor{hmLegend49}{HTML}{EEB683}
\fill[hmLegend49] (2.1881,0.290) rectangle (2.2068,0.390);
\definecolor{hmLegend50}{HTML}{EEB480}
\fill[hmLegend50] (2.2063,0.290) rectangle (2.2249,0.390);
\definecolor{hmLegend51}{HTML}{EDB27C}
\fill[hmLegend51] (2.2244,0.290) rectangle (2.2430,0.390);
\definecolor{hmLegend52}{HTML}{EDB079}
\fill[hmLegend52] (2.2425,0.290) rectangle (2.2611,0.390);
\definecolor{hmLegend53}{HTML}{EDAE75}
\fill[hmLegend53] (2.2606,0.290) rectangle (2.2793,0.390);
\definecolor{hmLegend54}{HTML}{ECAC72}
\fill[hmLegend54] (2.2788,0.290) rectangle (2.2974,0.390);
\definecolor{hmLegend55}{HTML}{ECAA6F}
\fill[hmLegend55] (2.2969,0.290) rectangle (2.3155,0.390);
\definecolor{hmLegend56}{HTML}{ECA86C}
\fill[hmLegend56] (2.3150,0.290) rectangle (2.3336,0.390);
\definecolor{hmLegend57}{HTML}{ECA669}
\fill[hmLegend57] (2.3331,0.290) rectangle (2.3518,0.390);
\definecolor{hmLegend58}{HTML}{EBA567}
\fill[hmLegend58] (2.3513,0.290) rectangle (2.3699,0.390);
\definecolor{hmLegend59}{HTML}{EBA364}
\fill[hmLegend59] (2.3694,0.290) rectangle (2.3880,0.390);
\definecolor{hmLegend60}{HTML}{EAA162}
\fill[hmLegend60] (2.3875,0.290) rectangle (2.4061,0.390);
\definecolor{hmLegend61}{HTML}{EA9F5F}
\fill[hmLegend61] (2.4056,0.290) rectangle (2.4243,0.390);
\definecolor{hmLegend62}{HTML}{EA9E5D}
\fill[hmLegend62] (2.4238,0.290) rectangle (2.4424,0.390);
\definecolor{hmLegend63}{HTML}{EA9C5A}
\fill[hmLegend63] (2.4419,0.290) rectangle (2.4605,0.390);
\definecolor{hmLegend64}{HTML}{E99A58}
\fill[hmLegend64] (2.4600,0.290) rectangle (2.4786,0.390);
\definecolor{hmLegend65}{HTML}{E99855}
\fill[hmLegend65] (2.4781,0.290) rectangle (2.4968,0.390);
\definecolor{hmLegend66}{HTML}{E89753}
\fill[hmLegend66] (2.4963,0.290) rectangle (2.5149,0.390);
\definecolor{hmLegend67}{HTML}{E89550}
\fill[hmLegend67] (2.5144,0.290) rectangle (2.5330,0.390);
\definecolor{hmLegend68}{HTML}{E8934E}
\fill[hmLegend68] (2.5325,0.290) rectangle (2.5511,0.390);
\definecolor{hmLegend69}{HTML}{E8914B}
\fill[hmLegend69] (2.5506,0.290) rectangle (2.5693,0.390);
\definecolor{hmLegend70}{HTML}{E79049}
\fill[hmLegend70] (2.5688,0.290) rectangle (2.5874,0.390);
\definecolor{hmLegend71}{HTML}{E78E46}
\fill[hmLegend71] (2.5869,0.290) rectangle (2.6055,0.390);
\definecolor{hmLegend72}{HTML}{E68C44}
\fill[hmLegend72] (2.6050,0.290) rectangle (2.6236,0.390);
\definecolor{hmLegend73}{HTML}{E68A41}
\fill[hmLegend73] (2.6231,0.290) rectangle (2.6418,0.390);
\definecolor{hmLegend74}{HTML}{E6893F}
\fill[hmLegend74] (2.6413,0.290) rectangle (2.6599,0.390);
\definecolor{hmLegend75}{HTML}{E6873C}
\fill[hmLegend75] (2.6594,0.290) rectangle (2.6780,0.390);
\definecolor{hmLegend76}{HTML}{E5853A}
\fill[hmLegend76] (2.6775,0.290) rectangle (2.6961,0.390);
\definecolor{hmLegend77}{HTML}{E58337}
\fill[hmLegend77] (2.6956,0.290) rectangle (2.7143,0.390);
\definecolor{hmLegend78}{HTML}{E48235}
\fill[hmLegend78] (2.7138,0.290) rectangle (2.7324,0.390);
\definecolor{hmLegend79}{HTML}{E48032}
\fill[hmLegend79] (2.7319,0.290) rectangle (2.7505,0.390);
\definecolor{hmLegend80}{HTML}{E47F31}
\fill[hmLegend80] (2.7500,0.290) rectangle (2.7686,0.390);
\definecolor{hmLegend81}{HTML}{E47E30}
\fill[hmLegend81] (2.7681,0.290) rectangle (2.7868,0.390);
\definecolor{hmLegend82}{HTML}{E47E2F}
\fill[hmLegend82] (2.7863,0.290) rectangle (2.8049,0.390);
\definecolor{hmLegend83}{HTML}{E37D2E}
\fill[hmLegend83] (2.8044,0.290) rectangle (2.8230,0.390);
\definecolor{hmLegend84}{HTML}{E37C2D}
\fill[hmLegend84] (2.8225,0.290) rectangle (2.8411,0.390);
\definecolor{hmLegend85}{HTML}{E37C2C}
\fill[hmLegend85] (2.8406,0.290) rectangle (2.8593,0.390);
\definecolor{hmLegend86}{HTML}{E37B2B}
\fill[hmLegend86] (2.8588,0.290) rectangle (2.8774,0.390);
\definecolor{hmLegend87}{HTML}{E37B2A}
\fill[hmLegend87] (2.8769,0.290) rectangle (2.8955,0.390);
\definecolor{hmLegend88}{HTML}{E27A2A}
\fill[hmLegend88] (2.8950,0.290) rectangle (2.9136,0.390);
\definecolor{hmLegend89}{HTML}{E27A29}
\fill[hmLegend89] (2.9131,0.290) rectangle (2.9318,0.390);
\definecolor{hmLegend90}{HTML}{E27928}
\fill[hmLegend90] (2.9313,0.290) rectangle (2.9499,0.390);
\definecolor{hmLegend91}{HTML}{E27927}
\fill[hmLegend91] (2.9494,0.290) rectangle (2.9680,0.390);
\definecolor{hmLegend92}{HTML}{E27826}
\fill[hmLegend92] (2.9675,0.290) rectangle (2.9861,0.390);
\definecolor{hmLegend93}{HTML}{E17725}
\fill[hmLegend93] (2.9856,0.290) rectangle (3.0043,0.390);
\definecolor{hmLegend94}{HTML}{E17724}
\fill[hmLegend94] (3.0038,0.290) rectangle (3.0224,0.390);
\definecolor{hmLegend95}{HTML}{E17623}
\fill[hmLegend95] (3.0219,0.290) rectangle (3.0405,0.390);
\definecolor{hmLegend96}{HTML}{E17623}
\fill[hmLegend96] (3.0400,0.290) rectangle (3.0586,0.390);
\definecolor{hmLegend97}{HTML}{E17522}
\fill[hmLegend97] (3.0581,0.290) rectangle (3.0768,0.390);
\definecolor{hmLegend98}{HTML}{E17521}
\fill[hmLegend98] (3.0763,0.290) rectangle (3.0949,0.390);
\definecolor{hmLegend99}{HTML}{E07420}
\fill[hmLegend99] (3.0944,0.290) rectangle (3.1130,0.390);
\definecolor{hmLegend100}{HTML}{E0731F}
\fill[hmLegend100] (3.1125,0.290) rectangle (3.1311,0.390);
\definecolor{hmLegend101}{HTML}{E0731E}
\fill[hmLegend101] (3.1306,0.290) rectangle (3.1493,0.390);
\definecolor{hmLegend102}{HTML}{E0721D}
\fill[hmLegend102] (3.1488,0.290) rectangle (3.1674,0.390);
\definecolor{hmLegend103}{HTML}{E0721C}
\fill[hmLegend103] (3.1669,0.290) rectangle (3.1855,0.390);
\definecolor{hmLegend104}{HTML}{DF711C}
\fill[hmLegend104] (3.1850,0.290) rectangle (3.2036,0.390);
\definecolor{hmLegend105}{HTML}{DF711B}
\fill[hmLegend105] (3.2031,0.290) rectangle (3.2218,0.390);
\definecolor{hmLegend106}{HTML}{DF701A}
\fill[hmLegend106] (3.2213,0.290) rectangle (3.2399,0.390);
\definecolor{hmLegend107}{HTML}{DF7019}
\fill[hmLegend107] (3.2394,0.290) rectangle (3.2580,0.390);
\definecolor{hmLegend108}{HTML}{DF6F18}
\fill[hmLegend108] (3.2575,0.290) rectangle (3.2761,0.390);
\definecolor{hmLegend109}{HTML}{DE6E17}
\fill[hmLegend109] (3.2756,0.290) rectangle (3.2943,0.390);
\definecolor{hmLegend110}{HTML}{DE6E16}
\fill[hmLegend110] (3.2938,0.290) rectangle (3.3124,0.390);
\definecolor{hmLegend111}{HTML}{DE6D15}
\fill[hmLegend111] (3.3119,0.290) rectangle (3.3305,0.390);
\definecolor{hmLegend112}{HTML}{DE6D15}
\fill[hmLegend112] (3.3300,0.290) rectangle (3.3486,0.390);
\definecolor{hmLegend113}{HTML}{DE6D14}
\fill[hmLegend113] (3.3481,0.290) rectangle (3.3668,0.390);
\definecolor{hmLegend114}{HTML}{DE6C14}
\fill[hmLegend114] (3.3662,0.290) rectangle (3.3849,0.390);
\definecolor{hmLegend115}{HTML}{DE6C14}
\fill[hmLegend115] (3.3844,0.290) rectangle (3.4030,0.390);
\definecolor{hmLegend116}{HTML}{DE6C13}
\fill[hmLegend116] (3.4025,0.290) rectangle (3.4211,0.390);
\definecolor{hmLegend117}{HTML}{DD6B13}
\fill[hmLegend117] (3.4206,0.290) rectangle (3.4393,0.390);
\definecolor{hmLegend118}{HTML}{DD6B12}
\fill[hmLegend118] (3.4388,0.290) rectangle (3.4574,0.390);
\definecolor{hmLegend119}{HTML}{DD6B12}
\fill[hmLegend119] (3.4569,0.290) rectangle (3.4755,0.390);
\definecolor{hmLegend120}{HTML}{DD6B12}
\fill[hmLegend120] (3.4750,0.290) rectangle (3.4936,0.390);
\definecolor{hmLegend121}{HTML}{DD6A11}
\fill[hmLegend121] (3.4931,0.290) rectangle (3.5118,0.390);
\definecolor{hmLegend122}{HTML}{DD6A11}
\fill[hmLegend122] (3.5113,0.290) rectangle (3.5299,0.390);
\definecolor{hmLegend123}{HTML}{DD6A10}
\fill[hmLegend123] (3.5294,0.290) rectangle (3.5480,0.390);
\definecolor{hmLegend124}{HTML}{DD6910}
\fill[hmLegend124] (3.5475,0.290) rectangle (3.5661,0.390);
\definecolor{hmLegend125}{HTML}{DD6910}
\fill[hmLegend125] (3.5656,0.290) rectangle (3.5843,0.390);
\definecolor{hmLegend126}{HTML}{DC690F}
\fill[hmLegend126] (3.5838,0.290) rectangle (3.6024,0.390);
\definecolor{hmLegend127}{HTML}{DC680F}
\fill[hmLegend127] (3.6019,0.290) rectangle (3.6205,0.390);
\definecolor{hmLegend128}{HTML}{DC680E}
\fill[hmLegend128] (3.6200,0.290) rectangle (3.6386,0.390);
\definecolor{hmLegend129}{HTML}{DC680E}
\fill[hmLegend129] (3.6381,0.290) rectangle (3.6568,0.390);
\definecolor{hmLegend130}{HTML}{DC680E}
\fill[hmLegend130] (3.6562,0.290) rectangle (3.6749,0.390);
\definecolor{hmLegend131}{HTML}{DC670D}
\fill[hmLegend131] (3.6744,0.290) rectangle (3.6930,0.390);
\definecolor{hmLegend132}{HTML}{DC670D}
\fill[hmLegend132] (3.6925,0.290) rectangle (3.7111,0.390);
\definecolor{hmLegend133}{HTML}{DC670C}
\fill[hmLegend133] (3.7106,0.290) rectangle (3.7292,0.390);
\definecolor{hmLegend134}{HTML}{DC660C}
\fill[hmLegend134] (3.7287,0.290) rectangle (3.7474,0.390);
\definecolor{hmLegend135}{HTML}{DC660C}
\fill[hmLegend135] (3.7469,0.290) rectangle (3.7655,0.390);
\definecolor{hmLegend136}{HTML}{DB660B}
\fill[hmLegend136] (3.7650,0.290) rectangle (3.7836,0.390);
\definecolor{hmLegend137}{HTML}{DB660B}
\fill[hmLegend137] (3.7831,0.290) rectangle (3.8018,0.390);
\definecolor{hmLegend138}{HTML}{DB650B}
\fill[hmLegend138] (3.8013,0.290) rectangle (3.8199,0.390);
\definecolor{hmLegend139}{HTML}{DB650A}
\fill[hmLegend139] (3.8194,0.290) rectangle (3.8380,0.390);
\definecolor{hmLegend140}{HTML}{DB650A}
\fill[hmLegend140] (3.8375,0.290) rectangle (3.8561,0.390);
\definecolor{hmLegend141}{HTML}{DB6409}
\fill[hmLegend141] (3.8556,0.290) rectangle (3.8743,0.390);
\definecolor{hmLegend142}{HTML}{DB6409}
\fill[hmLegend142] (3.8738,0.290) rectangle (3.8924,0.390);
\definecolor{hmLegend143}{HTML}{DB6409}
\fill[hmLegend143] (3.8919,0.290) rectangle (3.9105,0.390);
\definecolor{hmLegend144}{HTML}{DB6408}
\fill[hmLegend144] (3.9100,0.290) rectangle (3.9286,0.390);
\definecolor{hmLegend145}{HTML}{DB6308}
\fill[hmLegend145] (3.9281,0.290) rectangle (3.9468,0.390);
\definecolor{hmLegend146}{HTML}{DA6307}
\fill[hmLegend146] (3.9463,0.290) rectangle (3.9649,0.390);
\definecolor{hmLegend147}{HTML}{DA6307}
\fill[hmLegend147] (3.9644,0.290) rectangle (3.9830,0.390);
\definecolor{hmLegend148}{HTML}{DA6207}
\fill[hmLegend148] (3.9825,0.290) rectangle (4.0011,0.390);
\definecolor{hmLegend149}{HTML}{DA6206}
\fill[hmLegend149] (4.0006,0.290) rectangle (4.0193,0.390);
\definecolor{hmLegend150}{HTML}{DA6206}
\fill[hmLegend150] (4.0188,0.290) rectangle (4.0374,0.390);
\definecolor{hmLegend151}{HTML}{DA6105}
\fill[hmLegend151] (4.0369,0.290) rectangle (4.0555,0.390);
\definecolor{hmLegend152}{HTML}{DA6105}
\fill[hmLegend152] (4.0550,0.290) rectangle (4.0736,0.390);
\definecolor{hmLegend153}{HTML}{DA6105}
\fill[hmLegend153] (4.0731,0.290) rectangle (4.0918,0.390);
\definecolor{hmLegend154}{HTML}{DA6104}
\fill[hmLegend154] (4.0913,0.290) rectangle (4.1099,0.390);
\definecolor{hmLegend155}{HTML}{D96004}
\fill[hmLegend155] (4.1094,0.290) rectangle (4.1280,0.390);
\definecolor{hmLegend156}{HTML}{D96003}
\fill[hmLegend156] (4.1275,0.290) rectangle (4.1461,0.390);
\definecolor{hmLegend157}{HTML}{D96003}
\fill[hmLegend157] (4.1456,0.290) rectangle (4.1643,0.390);
\definecolor{hmLegend158}{HTML}{D95F03}
\fill[hmLegend158] (4.1638,0.290) rectangle (4.1824,0.390);
\definecolor{hmLegend159}{HTML}{D95F02}
\fill[hmLegend159] (4.1819,0.290) rectangle (4.2005,0.390);
\draw[black!22,line width=0.25pt] (1.3,0.290) rectangle (4.2,0.390);
\draw[black!45,line width=0.25pt] (1.300,0.290) -- (1.300,0.250);
\node[font=\fontsize{6.5}{7.1}\selectfont\bfseries] at (1.300,0.140) {-10};
\draw[black!45,line width=0.25pt] (1.590,0.290) -- (1.590,0.250);
\node[font=\fontsize{6.5}{7.1}\selectfont\bfseries] at (1.590,0.140) {0};
\draw[black!45,line width=0.25pt] (1.880,0.290) -- (1.880,0.250);
\node[font=\fontsize{6.5}{7.1}\selectfont\bfseries] at (1.880,0.140) {10};
\draw[black!45,line width=0.25pt] (2.460,0.290) -- (2.460,0.250);
\node[font=\fontsize{6.5}{7.1}\selectfont\bfseries] at (2.460,0.140) {30};
\draw[black!45,line width=0.25pt] (3.330,0.290) -- (3.330,0.250);
\node[font=\fontsize{6.5}{7.1}\selectfont\bfseries] at (3.330,0.140) {60};
\draw[black!45,line width=0.25pt] (4.200,0.290) -- (4.200,0.250);
\node[font=\fontsize{6.5}{7.1}\selectfont\bfseries] at (4.200,0.140) {90};
\end{tikzpicture}

%% file: sections/sec-6.tex
\section{Conclusion}

This paper introduced TreeGraft, a multi-drafter framework for tree-based speculative decoding organized around three questions: \emph{where} to graft, \emph{how} to graft, and \emph{when} to graft. Expanded grafting-position selection and non-destructive grafting address the first two questions by improving tree quality, while value-guided online scheduling addresses the third by controlling drafting cost. Together, these components turn shared-tree grafting into a controlled quality--cost trade-off. Experiments show that TreeGraft outperforms both single-drafter endpoints on average and closely tracks the better endpoint when one endpoint dominates, while ablations confirm that the components contribute complementary gains. We discuss limitations and future directions in Appendix~\ref{app:limitations}.

\newpage

\section*{Acknowledgments}

This work was supported by the National Natural Science Foundation of China (Grant No.~62301266), the National Natural Science Foundation of China (Grant No.~62576091), and the Jiangsu Province Carbon Peak Carbon Neutrality Science and Technology Innovation Special Fund Project (Grant No.~BT2025029). We also acknowledge support from the Southeast University Big Data Computing Center and the Southeast University Kunpeng \& Ascend Center of Cultivation.

\noindent\textbf{Competing Interests.} The authors declare no competing interests.

%% file: sections/appendix.tex
\appendix

\section{Limitations}
\label{app:limitations}

TreeGraft is drafter training-free and does not train additional draft heads or auxiliary draft modules. This makes it easy to apply with off-the-shelf draft sources, but also means that it does not replace training-based draft-source improvement methods, such as EAGLE-style draft heads~\citep{Li2025EAGLE3}. As discussed in Section~\ref{sec:related_work}, these methods are largely orthogonal to TreeGraft, since TreeGraft operates at the shared-tree construction and scheduling level. Combining TreeGraft with trained draft sources is therefore a natural direction for future work. 

In addition, because the smallest neural LLMs in our implementation still introduce nontrivial forwarding latency, we use an n-gram drafter as the low-cost small drafter. This lookup-based drafter is one practical instantiation rather than an exhaustive choice, and future work can explore stronger lightweight draft sources, such as Engram-style conditional-memory lookup~\citep{cheng2026engram} or suffix-tree-based drafters, within the same TreeGraft framework.

\section{Details of Online Scheduling}
\label{app:online_scheduling}

This appendix provides the technical implementation details, hyperparameters, network architectures, candidate signals, and offline data construction procedure for the online scheduling method introduced in Section~\ref{sec:online_scheduling}. 

\subsection{Fitting Data Collection}
\label{app:fitting_data}
The offline fitting data is collected by executing complete scheduling trajectories. We set the maximum number of drafting steps to \(D=5\). A complete scheduling trajectory is denoted by
\[
\tau_{1:D}\in\{0,1\}^D,
\]
where \(\tau_k=1\) means that the middle drafter is called at the \(k\)-th drafting step, and \(\tau_k=0\) means that the middle drafter is skipped.
\paragraph{Offline reports.}
The offline stage uses 6 training target--middle model pairs:
\[
\begin{gathered}
\text{LLaMA 3.3-70B-Instruct / LLaMA 3.2-1B-Instruct},\\
\text{LLaMA 3.3-70B-Instruct / LLaMA 3.2-3B-Instruct},\\
\text{LLaMA 3.1-8B-Instruct / LLaMA 3.2-1B-Instruct},\\
\text{Qwen3-32B / Qwen3-0.6B},\quad
\text{Qwen3-32B / Qwen3-8B},\quad
\text{Qwen3-32B / Qwen3-1.7B}.
\end{gathered}
\]
It uses 5 fitting datasets:
\[
\texttt{Alpaca\_fit},\quad
\texttt{GSM8K\_fit},\quad
\texttt{HumanEval\_fit},\quad
\texttt{NQ\_fit},\quad
\texttt{CNN/DM\_fit}.
\]
This gives \(6\times 5=30\) fitting reports. The online evaluation uses disjoint evaluation subsets for the first five datasets, and additionally includes MT-Bench as a fully held-out task. We also evaluate on 4 held-out target--middle model pairs that are not used in the offline stage:
\[
\begin{gathered}
\text{LLaMA 3.3-70B-Instruct / LLaMA 3.1-8B-Instruct},\\
\text{LLaMA 3.1-8B-Instruct / LLaMA 3.2-3B-Instruct},\\
\text{Qwen3-8B / Qwen3-0.6B},\quad
\text{Qwen3-32B / Qwen3-4B}.
\end{gathered}
\]
\paragraph{Model-family coverage.}
The offline value system and the distilled online scheduler are evaluated across two representative
mainstream open-source model families, LLaMA 3 and Qwen3, covering multiple target--middle
scale combinations. Importantly, the online evaluation includes both model pairs used during offline
fitting and held-out target--middle pairs that are never used for value-system fitting or scheduler
distillation. Together with the fully held-out MT-Bench evaluation, this setup suggests that the
learned scheduling behavior captures general cost--quality trade-off patterns in shared-tree
construction, rather than being tied to a single model size, model family, or benchmark.

\paragraph{Scheduling traces.}
For each fitting report, we enumerate the complete binary scheduling space. Since \(D=5\), this gives \(2^5=32\) complete scheduling trajectories per report and \(30\times 32=960\) trajectory-level records in total. Executing one complete scheduling trajectory yields a scheduling trace, which contains: (1) the binary scheduling trajectory \(\tau_{1:D}\), (2) the trajectory-level accepted length \(A(\tau_{1:D})\) and latency \(t(\tau_{1:D})\), and (3) the pre-action online states \(z_1,\ldots,z_D\) observed along the trajectory.
Collecting these trajectories for all six fitting model pairs costs approximately 160 A100 GPU-hours in total (about 20 hours of wall-clock time with eight A100 GPUs).

During execution, latency is measured as part of the trajectory outcome. In the saved logs, the same
information is stored through the accepted length \(A(\tau_{1:D})\) and throughput \(\eta(\tau_{1:D})\), where
throughput is reported in tokens per second. Since the saved latency labels are represented in
milliseconds, we reconstruct the trajectory-level latency as
\[
t_{\mathrm{ms}}(\tau_{1:D}) = 1000 \cdot \frac{A(\tau_{1:D})}{\eta(\tau_{1:D})}.
\]
Thus, the latency label used in the value system is a trajectory-level latency in milliseconds
reconstructed from the trajectory-level accepted length and throughput. This is only a unit convention:
the main-text definition \(\eta=A/t\) corresponds to using latency \(t\) in seconds.

At each step \(k\), we record the pre-action state \(z_k\), the current action \(a_k\), the future suffix \(\tau_{k+1:D}\), and the trajectory-level labels \(A(\tau_{1:D})\) and \(t(\tau_{1:D})\). This gives fitting samples of the form
\[
(z_k,\;a_k,\;\tau_{k+1:D},\;A,\;t).
\]
The labels \(A\) and \(t\) are always defined at the complete-trajectory level, rather than as local one-step quantities.

\paragraph{Online state and runtime context.}
Following Section~\ref{sec:online_scheduling}, the online state is written as
\[
z_k=(\mathbf{c},\;k,\;\tau_{1:k-1},\;\phi_k),
\]
where \(\mathbf{c}\) is the runtime context, \(k\) is the drafting-step index, \(\tau_{1:k-1}\) is the scheduling history, and \(\phi_k\) denotes pre-action tree signals computed from the current draft tree.

The runtime context \(\mathbf{c}\) contains four features:
\[
\mathbf{c}=
\bigl[
\log(1+\mathrm{prompt\ length}),\;
t_{\mathrm{tgt}},\;
t_{\mathrm{mid}},\;
\log(t_{\mathrm{tgt}}/t_{\mathrm{mid}})
\bigr].
\]
Here \(t_{\mathrm{tgt}}\) and \(t_{\mathrm{mid}}\) denote the per-call timing of the target model and the middle drafter, respectively. In offline fitting, \(t_{\mathrm{tgt}}\) and \(t_{\mathrm{mid}}\) are read from each fitting report using the all-call warm-up trajectory \(\tau_{1:D}^{\mathrm{warm}}=(1,1,1,1,1)\). They correspond to the recorded average per-call target forward time and middle drafter forward time. The prompt-length feature is computed as the model-family--dataset average prompt length and then transformed by \(\log(1+\cdot)\). This report-level averaging is used because the offline trajectory labels are also aggregated at the report level; during online evaluation, \(\log(1+\mathrm{prompt\ length})\) is computed from the actual input prompt.

The scheduling history is encoded as a 7-dimensional vector:
\[
h_k =
\left[
k/D,\;
(k-1)/D,\;
\bigl(\mathbf{1}[j<k\ \mathrm{and}\ a_j=1]\bigr)_{j=1}^{D}
\right].
\]
Since \(D=5\) is fixed in the main experiments, no additional padding is required. Positions that are not in the prefix are naturally set to zero by the condition \(j<k\).

\paragraph{Canonical online states.}
When extracting training samples from scheduling traces, we first construct canonical online states. Each canonical state is identified by the tuple
\[
(\mathrm{report\_id},\;k,\;\tau_{1:k-1}).
\]
Different complete scheduling trajectories may visit the same canonical state. We merge these repeated occurrences into a single canonical state. The tree signals of the merged state are averaged over its occurrences, and the state weight is set to
\[
w=\sqrt{\mathrm{count}},
\]
where \(\mathrm{count}\) is the number of occurrences of the canonical state in the corresponding report. In the main experiments, this procedure yields 930 canonical online states, which are later used as the teacher states for constructing margin targets.

Each complete scheduling trajectory contributes \(D=5\) action-conditioned return rows. Therefore, the 960 trajectory-level records yield
\[
30\times 32\times 5 = 4800
\]
action-conditioned rows. Among them, 2400 rows correspond to \(a_k=0\) and 2400 rows correspond to \(a_k=1\), forming the skip-return and call-return datasets, respectively.

\subsection{Candidate Tree Signals}
\label{app:online_observations}

Table~\ref{tab:candidate_signals} defines the ten candidate tree signals used for state selection. All signals are computed before the action at the \(k\)-th drafting step, and therefore correspond to pre-action observations of the current draft tree.

\paragraph{Notation.}
Let \(\mathcal{T}_k\) denote the shared draft tree before the \(k\)-th drafting step, and \(\mathcal{F}_k\) denote the current frontier. For a node \(v\), let \(d(v)\) be its depth, \(s(v)\) its cumulative path score, and \(\mathrm{child}(v)\) its children.
We construct a reselection preview frontier to measure structural change potential. The depth set of the current frontier is
\[
D_F=\{d(v):v\in\mathcal{F}_k\}.
\]
Given a lookback window \(L\), the candidate depth set is
\[
D_C= \bigcup_{d\in D_F} \{x:\max(0,d-L)\le x\le d\}.
\]
The corresponding candidate set is
\[
\mathcal{R}_k=\{v\in V(\mathcal{T}_k):d(v)\in D_C\}.
\]
We sort \(\mathcal{R}_k\) by cumulative path score \(s(v)\) and take the top-\(W\) nodes to form the preview frontier:
\[
\widehat{\mathcal{F}}_k=\mathrm{Top}_W(\mathcal{R}_k,s).
\]
The main experiments set \(L=5\). A pre-action state is kept only when all ten candidate tree signals are finite. If any candidate signal is missing or non-finite, the corresponding state is discarded.

\begin{table*}[t]
\centering
\small
\caption{Candidate tree signals used for state selection.}
\resizebox{\textwidth}{!}{
\begin{tabular}{lll}
\toprule
Signal & Definition & Meaning \\
\midrule
\(\mathrm{depth\_before}\)
& \(\displaystyle \frac{1}{|\mathcal{F}_k|} \sum_{v\in\mathcal{F}_k} d(v)\)
& Average depth of the current frontier \\
\(\mathrm{leaf\_ratio\_before}\)
& \(\displaystyle \frac{1}{|\mathcal{F}_k|} \sum_{v\in\mathcal{F}_k} \mathbf{1}[|\mathrm{child}(v)|=0]\)
& Ratio of leaf nodes in the current frontier \\
\(\mathrm{pre\_reselect\_depth\_delta}\)
& \(\displaystyle \frac{1}{|\widehat{\mathcal{F}}_k|} \sum_{v\in\widehat{\mathcal{F}}_k}d(v) - \frac{1}{|\mathcal{F}_k|} \sum_{v\in\mathcal{F}_k}d(v)\)
& Change in average frontier depth after reselection preview \\
\(\mathrm{pre\_reselect\_cands}\)
& \(\displaystyle |\mathcal{R}_k|\)
& Number of candidate nodes in the reselection preview \\
\(\mathrm{max\_depth\_before}\)
& \(\displaystyle \max_{v\in\mathcal{F}_k} d(v)\)
& Maximum depth of the current frontier \\
\(\mathrm{collapse\_debt}\)
& \(\displaystyle \max(0,\;-\mathrm{pre\_reselect\_depth\_delta})\)
& Amount by which the preview frontier becomes shallower \\
\(\mathrm{pre\_reselect\_past\_ratio}\)
& \(\displaystyle \frac{1}{|\widehat{\mathcal{F}}_k|} \sum_{v\in\widehat{\mathcal{F}}_k} \mathbf{1}[d(v)\notin D_F]\)
& Ratio of preview-frontier nodes from past depths \\
\(\mathrm{pre\_reselect\_overlap}\)
& \(\displaystyle \frac{|\widehat{\mathcal{F}}_k\cap \mathcal{F}_k|} {|\widehat{\mathcal{F}}_k|}\)
& Overlap between the current and preview frontiers \\
\(\mathrm{pre\_reselect\_score}\)
& \(\displaystyle \frac{1}{|\{v\in\widehat{\mathcal{F}}_k:s(v)\ \mathrm{finite}\}|} \sum_{\substack{v\in\widehat{\mathcal{F}}_k\\ s(v)\ \mathrm{finite}}} s(v)\)
& Average cumulative score of finite-score nodes in the preview frontier \\
\(\mathrm{finite\_score\_ratio\_before}\)
& \(\displaystyle \frac{1}{|\mathcal{F}_k|} \sum_{v\in\mathcal{F}_k} \mathbf{1}[s(v)\ \mathrm{finite}]\)
& Ratio of finite-score nodes in the current frontier \\
\bottomrule
\end{tabular}
}
\label{tab:candidate_signals}
\end{table*}

\paragraph{Selected signal subset.}
The final 6-dimensional signal subset identified in the main experiments is
\[
\begin{aligned}
B^\star=\{&
\mathrm{depth\_before},\;
\mathrm{leaf\_ratio\_before},\;
\mathrm{collapse\_debt},\\
&
\mathrm{pre\_reselect\_past\_ratio},\;
\mathrm{pre\_reselect\_overlap},\;
\mathrm{pre\_reselect\_score}
\}.
\end{aligned}
\]
Thus, the tree-state part of the final online scheduler consists of these six pre-action observations.

\subsection{Value System Architecture and Training}
\label{app:value_system_fitting}

We denote the selected signal subset by $B^\star$, and the resulting tree-signal vector at step $k$ by $\phi^\star_k$. For each candidate subset \(B \subseteq \Phi_k\), the value system \(f_{\theta}^B\) predicts the accepted length and log-latency of the complete scheduling trajectory, conditioned on the current online state, the current action, and the future suffix:
\[
f_{\theta}^B(z_k^B,\;a_k,\;\tau_{k+1:D}).
\]
The future suffix is explicitly included because \(z_k\) and \(a_k\) alone do not determine the final trajectory-level outcome.

\paragraph{Input encoding.}
For a candidate subset \(B\), the value-system input dimension is \(17+|B|\). The input consists of
\[
\underbrace{\mathbf{c}}_{4\ \mathrm{dims}},
\quad
\underbrace{h_k}_{7\ \mathrm{dims}},
\quad
\underbrace{\phi_k^B}_{|B|\ \mathrm{dims}},
\quad
\underbrace{u_{k+1:D}}_{6\ \mathrm{dims}}.
\]
Here \(\mathbf{c}\) is the 4-dimensional runtime context, \(h_k\) is the 7-dimensional scheduling-history encoding, \(\phi_k^B\) is the selected tree-signal subset, and \(u_{k+1:D}\) encodes the future suffix. The future suffix encoding is
\[
u_{k+1:D}
=
\left[
(D-k)/D,\;
\bigl(\mathbf{1}[j>k\ \mathrm{and}\ a_j=1]\bigr)_{j=1}^{D}
\right].
\]
Positions that are not in the suffix are set to zero by the condition \(j>k\). After subset selection, the final selected subset has \(|B^\star|=6\), so the final value-system input has \(17+6=23\) dimensions.

The current action \(a_k\) is not concatenated as an additional binary input. Instead, the value system uses action-conditioned heads, so the action information is represented by the identity of the head.

\paragraph{Architecture.}
For each candidate subset \(B\), the value system consists of four independent MLP heads:
\[
(A \mid a_k=0),\quad
(\log t \mid a_k=0),\quad
(A \mid a_k=1),\quad
(\log t \mid a_k=1).
\]
For the final selected subset \(B^\star\), each MLP uses the architecture
\[
\mathrm{Linear}(23,128)
\to
\mathrm{ReLU}
\to
\mathrm{Linear}(128,64)
\to
\mathrm{ReLU}
\to
\mathrm{Linear}(64,1),
\]
comprising 11{,}393 parameters per head and 45{,}572 parameters in total. Inputs and outputs are standardized to zero mean and unit variance. Each head computes its own input and output standardization statistics on the corresponding training data.

The latency head predicts log-latency in milliseconds. When computing predicted throughput, we first recover
\[
\hat{t}_{\mathrm{ms}} = \exp(\widehat{\log t_{\mathrm{ms}}}),
\]
and then compute
\[
\hat{\eta} = 1000 \cdot \frac{\hat{A}}{\hat{t}_{\mathrm{ms}}},
\]
where the factor \(1000\) converts millisecond latency to the token-per-second throughput used in evaluation. This is consistent with the main-text definition \(\eta=A/t\) when \(t\) is measured in seconds.

\paragraph{Training samples.}
For a fixed canonical state \(z_k\), current action \(a_k\), and future suffix \(\tau_{k+1:D}\), we recover the corresponding complete scheduling trajectory and read its trajectory-level labels \(A\) and \(t\). Therefore, the value-system fitting data is arranged as action-conditioned return rows rather than as report-level aggregate regression. Although transition-style rows can be constructed in implementation, the main experiments use only action-conditioned return datasets for system fitting.

\paragraph{Training objective.}
Each head is trained using weighted MSE:
\begin{equation}
\mathcal{L}_h
=
\frac{\sum_i w_i\bigl(f_h(x_i)-y_i\bigr)^2}{\sum_i w_i},
\label{eq:value_system_loss}
\end{equation}
where \(w_i\) is the canonical-state weight inherited from the merged state. Each value-system head is trained with AdamW using learning rate \(10^{-2}\), weight decay \(10^{-4}\), maximum 160 epochs, early-stopping patience 20, and minimum improvement \(10^{-4}\). The four heads use the same base seed with different offsets. If a head has at least 40 training samples, we further split a small validation subset from its training data for early stopping; otherwise, no additional validation split is used.

\paragraph{Tree-signal subset selection.}
Subset selection evaluates every non-empty subset of the 10 candidate tree signals, resulting in
\[
2^{10}-1=1023
\]
candidate bases. For each candidate subset \(B\), we fit a separate value system and evaluate it using five-fold grouped out-of-fold validation. The grouping unit is the fitting report: all trajectory rows and canonical states from the same report are assigned to the same fold. Reports are assigned to folds in a round-robin order, which prevents rows from the same report from appearing on both the training and validation sides.

For each candidate subset \(B\), we compute out-of-fold predictions on the skip-return and call-return datasets. On each action branch, we evaluate the normalized mean absolute error (NMAE) for accepted length, latency, and throughput. We define
\[
\mathrm{NMAE}(y,\widehat{y})
=
\frac{\frac{1}{n}\sum_{i=1}^{n}|y_i-\widehat{y}_i|}
{\mathrm{std}(y)}.
\]
The NMAE is computed on out-of-fold rows using ordinary MAE and ordinary standard deviation, without additional sample weights.

For the skip branch, we compute
\[
E_{\mathrm{skip}}(B)
=
\frac{1}{3}
\left(
\mathrm{NMAE}_{A,0}(B)
+
\mathrm{NMAE}_{t,0}(B)
+
\mathrm{NMAE}_{\eta,0}(B)
\right),
\]
and for the call branch,
\[
E_{\mathrm{call}}(B)
=
\frac{1}{3}
\left(
\mathrm{NMAE}_{A,1}(B)
+
\mathrm{NMAE}_{t,1}(B)
+
\mathrm{NMAE}_{\eta,1}(B)
\right).
\]
The identification score is
\[
\mathrm{id\_score}(B)
=
\frac{1}{2}
\left(
E_{\mathrm{skip}}(B)+E_{\mathrm{call}}(B)
\right).
\]
We select the 6-dimensional subset \(B^\star\) that minimizes this score. In the main experiments, the selected subset achieves
\[
\mathrm{id\_score}(B^\star)=0.1863.
\]

\subsection{Margin Predictor Architecture}
\label{app:runtime_policy}

The lightweight runtime scheduler \(g_\psi(z_k^\star)\) regresses the optimal action-value margin \(y_k\). After fitting the selected value system, we construct this margin by comparing the best predicted future throughput under call and skip:
\begin{equation}
y_k
=
\max_{\tau_{k+1:D}}
\widehat{\eta}_{\theta^\star}(z_k^\star,\;1,\;\tau_{k+1:D})
-
\max_{\tau_{k+1:D}}
\widehat{\eta}_{\theta^\star}(z_k^\star,\;0,\;\tau_{k+1:D}).
\label{eq:appendix_margin_target}
\end{equation}
Since \(D=5\), the maximization over future suffixes is implemented by exhaustive enumeration over all \(2^{D-k}\) binary suffixes. Thus, the margin target is produced by an exact suffix planner within the enumerated call/skip space, rather than by approximate search.

\paragraph{Architecture.}
The predictor \(g_\psi\) is a three-layer MLP:
\[
\mathrm{Linear}(17,64)
\to
\mathrm{ReLU}
\to
\mathrm{Linear}(64,32)
\to
\mathrm{ReLU}
\to
\mathrm{Linear}(32,1),
\]
with 3{,}265 parameters. Its 17-dimensional input is
\[
17 = 4 + 7 + 6,
\]
consisting of the 4-dimensional runtime context, the 7-dimensional scheduling-history encoding, and the selected 6-dimensional tree-signal subset. Unlike the value system, the margin predictor explicitly excludes the future suffix \(\tau_{k+1:D}\). The suffix-dependent reasoning is handled offline by the value system and exhaustive suffix maximization.

\paragraph{Training.}
After constructing margin teachers for the 930 canonical states, the margin predictor is trained using unweighted MSE:
\[
\mathcal{L}_{\mathrm{margin}}
=
\frac{1}{n}
\sum_{i=1}^{n}
\bigl(g_\psi(z_i^\star)-y_i\bigr)^2.
\]
The margin-predictor training objective does not use sample weights. However, when summarizing teacher quality, such as teacher MAE or sign accuracy, we use canonical-state weights for weighted reporting.

During online inference, the policy maps the sign of the predicted margin to the scheduling action:
\[
\pi_\psi(z_k^\star)
=
\begin{cases}
1\;(\text{call}), & g_\psi(z_k^\star)>0,\\
0\;(\text{skip}), & \text{otherwise}.
\end{cases}
\]
Thus, online scheduling only requires a single lightweight forward pass at each drafting step.

\subsection{Offline Evaluation Results}
\label{app:evaluation_results}

Table~\ref{tab:id_scheduler_result} summarizes the offline evaluation metrics for the final scheduler distillation on scheduling traces.

During the state selection phase, the selected 6-dimensional state minimizes the grouped out-of-fold identification score compared with using either the single best signal, which lacks sufficient context, or all 10 candidate signals, which introduces redundancy and noise.

Building upon this optimized state, we evaluate the offline scheduler solving result. As shown in Table~\ref{tab:id_scheduler_result}, the TreeGraft online scheduler achieves 24.30 tok/s on the scheduling traces, close to the best enumerated scheduling trajectory in the call/skip space, which achieves 24.78 tok/s. This demonstrates that the action-value margin constructed from the offline value system effectively supervises the lightweight online scheduler, enabling it to approximate the best scheduling trajectory within the enumerated call/skip space.

\begin{table}[h]
\centering
\small
\caption{Scheduler distillation evaluation on scheduling traces. The online scheduler closely approximates the highest throughput achieved by the enumerated scheduling trajectories in the call/skip space.}
\begin{tabular}{lc}
\toprule
Method & Throughput (tok/s) \\
\midrule
Best enumerated scheduling trajectory in the call/skip space & \textbf{24.78} \\
TreeGraft online scheduler & 24.30 \\
\bottomrule
\end{tabular}
\label{tab:id_scheduler_result}
\end{table}

\subsection{Ablation on the Value System}
\label{app:value-system-ablation}

This appendix analyzes why TreeGraft does not directly deploy the fitted value system as the online scheduler. We compare the TreeGraft scheduler used in the main experiments with a direct value-system planner.

\paragraph{Direct value-system planner.}
The direct planner uses the fitted value system online. At each drafting step, it compares the predicted throughput of \emph{call} and \emph{skip}, after optimizing over future suffixes, and chooses the action with the higher predicted value. This is the most direct way to use the value system, but it also relies on accurate absolute predictions of accepted length and latency during online decoding.

\paragraph{Offline mean actual gap.}
We first define the offline metric. For each fitting report, let \(\eta_{\text{chosen}}\) denote the true throughput of the trajectory selected by a method, and let \(\eta_{\text{oracle}}\) denote the highest true throughput among the enumerated trajectories in the call/skip space of that report. This oracle is not a theoretical global optimum, but the best trajectory within the enumerated call/skip space. The actual gap on this report is defined as
\[
    \eta_{\text{oracle}} - \eta_{\text{chosen}} .
\]
The offline mean actual gap is the average actual gap over all fitting reports. A smaller value means that the method is closer to the enumerated oracle on the offline scheduling traces. However, this metric is not equivalent to the speedup obtained in real online evaluation.

\begin{table}[t]
\centering
\caption{Ablation on direct value-system planning. A smaller offline mean actual gap indicates a closer match to the offline enumerated oracle. Overall speedup is measured in real online evaluation relative to autoregressive target decoding.}
\label{tab:value-system-ablation}
\small
\begin{tabular}{lcc}
\toprule
Method & Offline mean actual gap & Overall speedup \\
\midrule
Direct value-system planner & 0.3718 & 0.95$\times$ \\
TreeGraft scheduler & 0.4826 & \textbf{1.60$\times$} \\
\bottomrule
\end{tabular}
\end{table}

\paragraph{Results.}
Table~\ref{tab:value-system-ablation} shows that a smaller offline gap does not necessarily translate into better online speedup. The direct value-system planner achieves a smaller offline gap than the TreeGraft scheduler, but its online speedup drops to only \(0.95\times\). This indicates that directly using the fitted value system for online planning is unreliable, because small prediction errors in accepted length and latency can be amplified when the planner optimizes over future suffixes.

TreeGraft therefore uses the value system only as an offline teacher. It first identifies useful tree signals and then converts trajectory-level accepted length and latency outcomes into a relative call/skip margin. The lightweight runtime scheduler is trained on this margin and only uses the current online state during inference. This design avoids deploying the future-conditioned value system directly online, while achieving substantially better end-to-end speedup.

\section{Runtime Analysis}
\label{app:runtime}

\subsection{Small Drafter Implementation and Cost Analysis}
\label{app:ngram_small_drafter}

In TreeGraft, the \emph{small drafter} is the low-cost proposal source used when the scheduler skips the middle drafter. In our implementation, it is a training-free n-gram drafter rather than a neural LLM. It performs no neural-network forward pass; instead, it retrieves candidate tokens and proposal scores through n-gram lookup over the available context, serving as the small drafter expansion step in Section~\ref{sec:grafting}.

During prefill, we cache the top-\(m\) candidate tokens at each context position together with their target-model probabilities from the prefill logits. Concretely, for a context position \(i\), the target prefill provides a next-token distribution \(p_{\mathrm{tar}}(\cdot \mid x_{\le i})\); we store its top-\(m\) tokens and the corresponding probabilities as the cached proposal candidates and scores for that position. We also build an n-gram index over the context. At drafting step \(k\), the small drafter matches the suffix of the current prefix to the context using the longest available n-gram, retrieves the cached candidates at the matched position, and uses the cached probabilities as the edge scores \(p(u)\) for expansion. If no match is found, it gradually shortens the n-gram and finally falls back to the cached candidates associated with the position of the most frequent token in the prompt. These cached scores come only from target-model outputs already available during prefill, so the small drafter introduces no additional target forward pass and does not use future verification outcomes or post-prefill target logits. Its lookup cost is included in the reported runtime.

The use of an n-gram drafter is motivated by the latency of real small LLMs. Table~\ref{tab:per_call_latency_online} reports the per-call latency of representative neural drafters and target models under the online evaluation setting. Even the smallest neural model in each family incurs non-negligible forwarding cost: LLaMA 3.2-1B-Instruct requires \(29.1\) ms per call, and Qwen3-0.6B requires \(44.8\) ms per call. Therefore, using a neural model as the small drafter would make the low-cost endpoint substantially more expensive.

\begin{table}[t]
\centering
\small
\setlength{\tabcolsep}{8pt}
\caption{Per-call latency of representative neural drafters and target models under the online evaluation setting in Section~\ref{sec:experiments}.}
\label{tab:per_call_latency_online}
\begin{tabular}{llr}
\toprule
\textbf{Model family} & \textbf{Model} & \textbf{Per-call latency (ms)} \\
\midrule
LLaMA & LLaMA 3.3-70B-Instruct target & \(122.2\) \\
LLaMA & LLaMA 3.2-1B-Instruct & \(29.1\) \\
LLaMA & LLaMA 3.1-8B-Instruct & \(48.9\) \\
\midrule
Qwen & Qwen3-32B target & \(77.6\) \\
Qwen & Qwen3-0.6B & \(44.8\) \\
Qwen & Qwen3-8B & \(55.0\) \\
\bottomrule
\end{tabular}
\end{table}

Table~\ref{tab:small_drafter_latency} directly compares the candidate small drafter choices. The n-gram drafter costs only \(0.92\) ms per call, while the smallest neural alternatives are tens of milliseconds per call. This latency gap is the reason we use the n-gram drafter to define the low-cost endpoint in the main experiments.

\begin{table}[t]
\centering
\small
\setlength{\tabcolsep}{8pt}
\caption{Per-call latency of candidate small drafter choices under the online evaluation setting in Section~\ref{sec:experiments}.}
\label{tab:small_drafter_latency}
\begin{tabular}{llr}
\toprule
\textbf{Small drafter} & \textbf{Type} & \textbf{Per-call latency (ms)} \\
\midrule
N-gram (ours) & Lookup-based & \(0.92\) \\
LLaMA 3.2-1B-Instruct & Neural LLM & \(29.1\) \\
Qwen3-0.6B & Neural LLM & \(44.8\) \\
\bottomrule
\end{tabular}
\end{table}

Therefore, the goal of All Small is not to represent the strongest possible small drafter, but to provide a genuinely low-cost endpoint for tree construction. This endpoint makes the scheduler's role clear: it decides when a real middle drafter call is worth paying for, rather than comparing two already expensive neural proposal sources.

TreeGraft itself is not restricted to n-gram small drafters. In principle, the same shared-tree grafting and scheduling framework can be combined with stronger neural small drafters or stronger lightweight lookup-based proposers. However, any such replacement requires a separate cost--quality study. We leave this direction for future work.

\subsection{Online Scheduler Cost Analysis}
\label{app:scheduler_runtime}

The online scheduler is called once per drafting step, so its overhead must be small enough not to become a runtime bottleneck. Table~\ref{tab:scheduler_cost} compares the scheduler latency with the main runtime components under the online evaluation setting.

\begin{table}[t]
\centering
\small
\setlength{\tabcolsep}{8pt}
\caption{Per-call latency of the online scheduler and the main runtime components under the online evaluation setting in Section~\ref{sec:experiments}.}
\label{tab:scheduler_cost}
\begin{tabular}{lrr}
\toprule
\textbf{Component} & \textbf{Per-call latency (ms)} & \textbf{Ratio vs.\ target} \\
\midrule
Target-side verification & \(69.39\) & \(1.000\times\) \\
Middle-model correction  & \(44.23\) & \(0.637\times\) \\
N-gram small drafter     & \(0.92\)  & \(0.013\times\) \\
Online scheduler         & \(0.318\) & \(0.005\times\) \\
\bottomrule
\end{tabular}
\end{table}

The scheduler costs only \(0.318\) ms per call, which is lower than the n-gram lookup cost and far smaller than neural forwarding components. With \(D=5\), its total cost per decoding turn is at most \(5\times0.318\approx1.59\) ms. Therefore, the scheduler is not a runtime bottleneck; the end-to-end speedup mainly comes from selectively avoiding unnecessary middle drafter calls rather than from ignoring scheduling overhead.

\section{Online Evaluation Setting Details}
\label{app:online_eval_setting_details}

The formal online evaluation is conducted on the 10 target--middle model pairs reported in Table~\ref{tab:main_results}. The blue rows in Table~\ref{tab:main_results} correspond to held-out model pairs, which are not used during offline state selection, value-system fitting, or scheduler distillation. The green MT-Bench column corresponds to held-out task evaluation, since no MT-Bench samples are used in the offline stage. The shortened LLaMA names in Table~\ref{tab:main_results} refer to instruct checkpoints: 70B denotes LLaMA 3.3-70B-Instruct, 8B denotes LLaMA 3.1-8B-Instruct, 3B denotes LLaMA 3.2-3B-Instruct, and 1B denotes LLaMA 3.2-1B-Instruct. For Qwen pairs, all models are from the Qwen3 family.

All online evaluations are conducted on two NVIDIA A100 GPUs with batch size \(1\). All compared methods use the same hardware, batching, tree-construction budget, decoding parameters, and target-verification procedure. For each benchmark, we randomly sample \(80\) evaluation examples, and the same sampled examples are used for TreeGraft, All Small, All Mid, and all ablation variants.

All online evaluations use the same tree-construction hyperparameters and verification budget. We set the maximum number of drafting steps to \(D=5\), the expansion frontier size to \(W=10\), the number of child tokens generated per expanded node to \(m=10\), and the verification-tree budget to \(B_{\mathrm{ver}}=63\). Here, \(B_{\mathrm{ver}}\) denotes the maximum number of non-root draft nodes retained after final pruning and submitted to the target model for tree verification. We also set \(\textit{max\_new\_tokens}=256\) and \(\textit{temperature}=0\). In the main online evaluation, we do not impose a fixed middle drafter schedule; instead, the trained scheduler decides at each drafting step whether to call the middle drafter.

The runtime context \(\mathbf{c}\) contains four features: the logarithm of the prompt length, \(t_{\mathrm{tgt}}\), \(t_{\mathrm{mid}}\), and \(\log(t_{\mathrm{tgt}}/t_{\mathrm{mid}})\). The prompt length is directly computed from the input sample. The timing features \(t_{\mathrm{tgt}}\) and \(t_{\mathrm{mid}}\) denote per-step average latency and are measured during online warm-up. Specifically, for each model pair and benchmark, we run three warm-up rounds on the first evaluation sample with middle drafter calls forced at all five drafting steps, and use the average timing from the third warm-up round to compute \(t_{\mathrm{tgt}}\), \(t_{\mathrm{mid}}\), and their log ratio. The warm-up sample is used only to measure timing features; its accepted length, verification outcome, throughput, and scheduler decision are not used as supervision.

\section{Algorithmic Details of Tree Construction}
\label{app:tree-construction-algorithms}

This appendix gives the step-by-step pseudocode corresponding to the tree-construction procedure in Section~\ref{sec:grafting}. Algorithm~\ref{alg:standard_tree_construction} summarizes the standard single-drafter tree construction rule, and Algorithm~\ref{alg:graft_tree_construction} summarizes the TreeGraft shared-tree construction rule. Both algorithms use the same tree budget and the same target-side verification procedure; the difference lies only in how the draft tree is constructed before verification. Algorithm~\ref{alg:global_prune} gives the final global pruning procedure used by both methods.

\paragraph{Notation.}
Let \(r\) denote the root node representing the current verified prefix. At drafting step \(k\), let \(\mathcal{T}_k\) be the current draft tree and \(V(\mathcal{T}_k)\) its node set. Let \(\mathcal{N}_k\) denote the nodes newly generated at the previous step, with \(\mathcal{N}_1=\{r\}\). For each non-root node \(v\), \(p(v)\) denotes the stored edge score of the token at \(v\), and \(s(v)\) denotes its cumulative path score. The operation \(\mathrm{Top}_W(\mathcal{C},s)\) returns the \(W\) highest-scoring nodes in candidate pool \(\mathcal{C}\), and \(\mathrm{Top}_m(\ell(v))\) returns the top-\(m\) child tokens from the next-token logits \(\ell(v)\) under the prefix ending at \(v\). The final operation \(\mathrm{GlobalPrune}(\mathcal{T},B_{\mathrm{ver}})\), detailed in Algorithm~\ref{alg:global_prune}, prunes the constructed tree to the fixed verification budget \(B_{\mathrm{ver}}\) and returns the pruned verification tree \(\mathcal{T}_{\mathrm{verify}}\) together with the verification buffers.

\paragraph{Global score-based pruning.}
Both Algorithm~\ref{alg:standard_tree_construction} and Algorithm~\ref{alg:graft_tree_construction} call the same final pruning procedure before target verification. This step ranks non-root nodes in the constructed draft tree by cumulative path score \(s(v)\), using depth \(d(v)\) and node id \(\mathrm{id}(v)\) only to break ties. It then greedily keeps high-scoring nodes together with the ancestor chains needed to connect them to the root \(r\). The resulting \(\mathcal{T}_{\mathrm{verify}}\) satisfies the verification budget and remains a valid tree for building the tree mask, position ids, and retrieval paths.

\begin{algorithm}[H]
\caption{Standard single-drafter tree construction}
\label{alg:standard_tree_construction}
\begin{algorithmic}[1]
\Require Verified prefix, single drafter \(q\), target model \(p_{\mathrm{tar}}\), maximum drafting steps \(D\), frontier size \(W\), branching size \(m\), verification budget \(B_{\mathrm{ver}}\)
\Ensure Target-verified output tokens
\State Initialize the draft tree with root \(r\): \(\mathcal{T}_1 \leftarrow \{r\}\)
\State Initialize newly generated nodes: \(\mathcal{N}_1 \leftarrow \{r\}\)
\State Set root score \(s(r)\leftarrow 1\)
\For{\(k=1,\ldots,D\)}
    \State Form the candidate pool from newly generated nodes: \(\mathcal{C}_k \leftarrow \mathcal{N}_k\)
    \State Select the expansion frontier: \(\mathcal{F}_k \leftarrow \mathrm{Top}_W(\mathcal{C}_k,s)\)
    \State Initialize the newly generated set for the next step: \(\mathcal{N}_{k+1}\leftarrow \emptyset\)
    \For{each node \(v\in\mathcal{F}_k\)}
        \State Run the drafter on the prefix ending at \(v\), and obtain next-token logits \(\ell_q(v)\)
        \State Select child tokens: \(\mathcal{Y}_v \leftarrow \mathrm{Top}_m(\ell_q(v))\)
        \For{each token \(y\in\mathcal{Y}_v\)}
            \State Create a child node \(u=(v,y)\) under parent \(v\)
            \State Store the edge score \(p(u)\leftarrow q(y\mid \pi(v))\)
            \State Compute the cumulative path score \(s(u)\leftarrow s(v)\cdot p(u)\)
            \State Attach \(u\) to \(\mathcal{T}_k\) and add it to \(\mathcal{N}_{k+1}\)
        \EndFor
    \EndFor
    \State Set \(\mathcal{T}_{k+1}\leftarrow \mathcal{T}_k\)
\EndFor
\State Apply final global pruning: \(\mathcal{T}_{\mathrm{verify}}\leftarrow \mathrm{GlobalPrune}(\mathcal{T}_{D+1},B_{\mathrm{ver}})\)
\State Verify \(\mathcal{T}_{\mathrm{verify}}\) with \(p_{\mathrm{tar}}\) using tree attention and the standard rejection-sampling rule
\State \Return Target-verified output tokens
\end{algorithmic}
\end{algorithm}

\begin{algorithm}[H]
\caption{TreeGraft shared-tree construction with grafting}
\label{alg:graft_tree_construction}
\begin{algorithmic}[1]
\Require Verified prefix, small drafter \(q_{\mathrm{s}}\), middle drafter \(q_{\mathrm{m}}\), target model \(p_{\mathrm{tar}}\), schedule actions \(a_{1:D}\), maximum drafting steps \(D\), frontier size \(W\), branching size \(m\), verification budget \(B_{\mathrm{ver}}\)
\Ensure Target-verified output tokens
\State Initialize the shared draft tree with root \(r\): \(\mathcal{T}_1\leftarrow \{r\}\)
\State Initialize newly generated nodes: \(\mathcal{N}_1\leftarrow \{r\}\)
\State Initialize the set of nodes visited by the middle drafter: \(\mathcal{P}_1\leftarrow \emptyset\)
\State Set root score \(s(r)\leftarrow 1\)
\For{\(k=1,\ldots,D\)}
    \If{\(a_k=0\)} \Comment{small drafter step}
        \State Form the candidate pool using the standard rule: \(\mathcal{C}_k\leftarrow \mathcal{N}_k\)
        \State Reuse the stored edge scores \(p(\cdot)\) and cumulative scores \(s(\cdot)\)
        \State Select the expansion frontier: \(\mathcal{F}_k\leftarrow \mathrm{Top}_W(\mathcal{C}_k,s)\)
        \State Initialize the newly generated set: \(\mathcal{N}_{k+1}\leftarrow \emptyset\)
        \For{each node \(v\in\mathcal{F}_k\)}
            \State Run the small drafter on the prefix ending at \(v\), and obtain next-token logits \(\ell_{\mathrm{s}}(v)\)
            \State Select child tokens: \(\mathcal{Y}_v\leftarrow \mathrm{Top}_m(\ell_{\mathrm{s}}(v))\)
            \For{each token \(y\in\mathcal{Y}_v\)}
                \State Create a child node \(u=(v,y)\) under parent \(v\)
                \State Store the edge score \(p(u)\leftarrow q_{\mathrm{s}}(y\mid \pi(v))\)
                \State Compute the cumulative path score \(s(u)\leftarrow s(v)\cdot p(u)\)
                \State Attach \(u\) to \(\mathcal{T}_k\) and add it to \(\mathcal{N}_{k+1}\)
            \EndFor
        \EndFor
        \State Keep the middle drafter visited set unchanged: \(\mathcal{P}_{k+1}\leftarrow \mathcal{P}_k\)
    \Else \Comment{middle drafter step}
        \State Form the expanded grafting-position candidate pool: \(\mathcal{C}_k\leftarrow V(\mathcal{T}_k)\setminus \mathcal{P}_k\)
        \State Run one middle drafter forward pass over \(\mathcal{C}_k\)
        \State Use this forward pass to update candidate edge scores \(p(v)\) for each non-root \(v\in\mathcal{C}_k\), keep \(s(r)=1\), recompute cumulative scores \(s(v)\), and cache the next-token logits \(\ell_{\mathrm{m}}(v)\)
        \State Select the grafting frontier using the updated scores: \(\mathcal{F}_k\leftarrow \mathrm{Top}_W(\mathcal{C}_k,s)\)
        \State Initialize the newly generated set: \(\mathcal{N}_{k+1}\leftarrow \emptyset\)
        \For{each node \(v\in\mathcal{F}_k\)}
            \State Reuse the cached middle drafter logits \(\ell_{\mathrm{m}}(v)\)
            \State Select child tokens: \(\mathcal{Y}_v\leftarrow \mathrm{Top}_m(\ell_{\mathrm{m}}(v))\)
            \For{each token \(y\in\mathcal{Y}_v\)}
                \State Create a child node \(u=(v,y)\) under parent \(v\)
                \State Store the edge score \(p(u)\leftarrow q_{\mathrm{m}}(y\mid \pi(v))\)
                \State Compute the cumulative path score \(s(u)\leftarrow s(v)\cdot p(u)\)
                \State Attach \(u\) under \(v\) without overwriting any existing child or branch
                \State Add \(u\) to \(\mathcal{N}_{k+1}\)
            \EndFor
        \EndFor
        \State Mark all nodes in the middle drafter candidate pool as visited: \(\mathcal{P}_{k+1}\leftarrow \mathcal{P}_k\cup \mathcal{C}_k\)
    \EndIf
    \State Set \(\mathcal{T}_{k+1}\leftarrow \mathcal{T}_k\)
\EndFor
\State Apply final global pruning: \(\mathcal{T}_{\mathrm{verify}}\leftarrow \mathrm{GlobalPrune}(\mathcal{T}_{D+1},B_{\mathrm{ver}})\)
\State Verify \(\mathcal{T}_{\mathrm{verify}}\) with \(p_{\mathrm{tar}}\) using tree attention and the standard rejection-sampling rule
\State \Return Target-verified output tokens
\end{algorithmic}
\end{algorithm}

\begin{algorithm}[H]
\caption{Global score-based pruning with ancestor closure}
\label{alg:global_prune}
\begin{algorithmic}[1]
\Require Draft tree \(\mathcal{T}\) with root \(r\), node scores \(s(v)\), verification budget \(B_{\mathrm{ver}}\)
\Ensure Pruned verification tree \(\mathcal{T}_{\mathrm{verify}}\), tree mask, position ids, and retrieval paths
\State Let \(M \leftarrow B_{\mathrm{ver}}+1\), where the extra node is the root
\If{\(|V(\mathcal{T})|\le M\)}
    \State \Return \(\mathcal{T}\) and its verification buffers
\EndIf
\State Compute the depth \(d(v)\) of every node \(v\in V(\mathcal{T})\)
\State Rank all non-root nodes by decreasing score:
\[
v_1,v_2,\ldots
\leftarrow
\mathrm{Sort}\bigl(V(\mathcal{T})\setminus\{r\},
(-s(v), d(v), \mathrm{id}(v))\bigr)
\]
\State Initialize the kept node set: \(\mathcal{K}\leftarrow \{r\}\)
\For{each ranked node \(v_i\)}
    \State Trace the ancestor chain from \(v_i\) upward until reaching a node already in \(\mathcal{K}\):
    \[
    \mathcal{A}(v_i) \leftarrow \{v_i,\mathrm{parent}(v_i),\ldots\}\setminus \mathcal{K}
    \]
    \If{\(|\mathcal{K}|+|\mathcal{A}(v_i)| \le M\)}
        \State Add the whole chain: \(\mathcal{K}\leftarrow \mathcal{K}\cup \mathcal{A}(v_i)\)
    \EndIf
    \If{\(|\mathcal{K}|=M\)}
        \State \textbf{break}
    \EndIf
\EndFor
\State Construct the induced subtree \(\mathcal{T}_{\mathrm{verify}}\leftarrow \mathcal{T}[\mathcal{K}]\)
\State Re-index the kept nodes so that parents always appear before children
\State Build the tree-attention mask:
\[
\mathrm{mask}(u,v)=1
\quad\text{iff}\quad
v=r \;\text{or}\; v \text{ is an ancestor of } u \;\text{or}\; v=u
\]
\State Set each node's position id to its depth in \(\mathcal{T}_{\mathrm{verify}}\)
\State Extract all root-to-leaf paths in \(\mathcal{T}_{\mathrm{verify}}\) as retrieval paths and pad shorter paths with \(-1\)
\State \Return \(\mathcal{T}_{\mathrm{verify}}\), tree mask, position ids, and retrieval paths
\end{algorithmic}
\end{algorithm}

\section{Qualitative Examples of Non-destructive Grafting}
\label{app:nd-grafting-examples}

To complement the quantitative ablation in Table~\ref{tab:ablation}, we provide two traced examples
showing why non-destructive grafting is needed. The trace collection follows the online evaluation
setting in Section~\ref{sec:experiments}. We use the Qwen3-32B/Qwen3-0.6B target--middle pair
on CNN/DailyMail, with the same tree-construction budget and decoding configuration as in
Section~\ref{sec:experiments}. To isolate the grafting mechanism, we use the fixed scheduling
pattern $[1,0,1,0,1]$, where 1 denotes calling the middle drafter and 0 denotes using the small
drafter. Thus, the middle drafter is called at drafting steps 1, 3, and 5.

Each example focuses on one middle drafter step. Under expanded grafting-position selection, the
middle drafter may select a historical node, i.e., a node that already exists in the shared tree before
the current middle drafter step. This selected historical node may already have children and a subtree.
Non-destructive grafting attaches the newly generated middle drafter children under this node without
removing the existing children. The examples below show cases where the existing subtree contains
a branch later accepted by the target model. If overwriting were used instead, that accepted branch
would be discarded.

We use the following notation:

\histtok{[H]}: node already present before the current middle drafter step,

\midtok{[M]}: node newly attached by the current middle drafter step.

In the text trees, \tok{*} marks the selected grafting position, and \tok{+} marks nodes on the later
target-accepted path. Omitted branches are denoted by \tok{...}.

\paragraph{Case 1:}
The selected grafting position is the historical node \histtok{.}. Before the current middle drafter step,
this node already contains the historical branch \histtok{L -> umber -> Liquid -> ators}. The target later
accepts this branch, forming the continuation \histtok{. Lumber Liquidators}.

\begin{center}
\begin{minipage}{0.70\linewidth}
\begin{TokenTree}
@histtok{[H] <root>}
\-- @histtok{[H] .} @tok{* +}
    |-- @histtok{[H] L} @tok{+}
    |   \-- @histtok{[H] umber} @tok{+}
    |       \-- @histtok{[H] Liquid} @tok{+}
    |           \-- @histtok{[H] ators} @tok{+}
    |-- @midtok{[M] The}
    |-- @midtok{[M] They}
    |-- @midtok{[M] CBS}
    \-- @tok{...}
\end{TokenTree}
\end{minipage}
\end{center}

Non-destructive grafting preserves the historical branch while attaching new middle drafter children
such as \midtok{The}, \midtok{They}, and \midtok{CBS}. With overwriting, the accepted continuation
\histtok{. Lumber Liquidators} would be removed.

\paragraph{Case 2:}
The selected grafting position is the historical node \histtok{that}. Before the current middle drafter
step, this node already contains the historical branch \histtok{P -> elle -> gr -> ini}. The target later
accepts this branch, forming the continuation \histtok{mentions that Pellegrini}.

\begin{center}
\begin{minipage}{0.70\linewidth}
\begin{TokenTree}
@histtok{[H] <root>}
\-- @histtok{[H] mentions} @tok{+}
    \-- @histtok{[H] that} @tok{* +}
        |-- @histtok{[H] P} @tok{+}
        |   \-- @histtok{[H] elle} @tok{+}
        |       \-- @histtok{[H] gr} @tok{+}
        |           \-- @histtok{[H] ini} @tok{+}
        |-- @midtok{[M] Martinez}
        |-- @midtok{[M] Hart}
        |-- @midtok{[M] the}
        \-- @tok{...}
\end{TokenTree}
\end{minipage}
\end{center}

Again, non-destructive grafting keeps the existing historical subtree while adding new middle drafter
children under the selected historical node. With overwriting, the target-accepted continuation
\histtok{mentions that Pellegrini} would be discarded.

\section{Additional Experimental Results}
\label{app:additional-results}

\subsection{Supplementary Study on the Small Drafter Choice}
\label{app:small drafter-choice}

In the main experiments, we use a training-free n-gram drafter as the small drafter.
This is because TreeGraft relies on the small drafter as a genuinely low-cost expansion source.
If the small drafter is replaced by a neural model with nontrivial forwarding cost, the default expansion steps may substantially reduce the overall speedup.

To examine this effect, we conduct a fixed-schedule preliminary study.
We remove the learned scheduler and call the middle drafter at drafting steps $1$, $3$, and $5$, i.e., using the schedule $[1,0,1,0,1]$.
For each middle--target pair, we compare two small drafter choices: the smallest pretrained neural model from the same family, and the training-free n-gram drafter.
All other settings, including the benchmark, tree budget, decoding hyperparameters, and middle--target pair, are kept unchanged.
All speedups are measured relative to autoregressive decoding with the target model.

\begin{table}[t]
\centering
\caption{Effect of the small drafter choice under a fixed middle drafter schedule.
The middle drafter is called at steps $1$, $3$, and $5$.
Values are speedups over autoregressive target decoding.}
\label{tab:small drafter-choice}
\small
\begin{tabularx}{\linewidth}{Xc}
\toprule
Small / Middle / Target & Speedup \\
\midrule
LLaMA 3.2-1B-Instruct / LLaMA 3.1-8B-Instruct / LLaMA 3.3-70B-Instruct
& 1.116$\times$ \\
n-gram / LLaMA 3.1-8B-Instruct / LLaMA 3.3-70B-Instruct
& 1.299$\times$ \\
\midrule
Qwen3-0.6B / Qwen3-4B / Qwen3-32B
& 0.735$\times$ \\
n-gram / Qwen3-4B / Qwen3-32B
& 1.108$\times$ \\
\midrule
Qwen3-0.6B / Qwen3-8B / Qwen3-32B
& 0.743$\times$ \\
n-gram / Qwen3-8B / Qwen3-32B
& 1.134$\times$ \\
\midrule
Average with neural small drafter
& 0.865$\times$ \\
Average with n-gram small drafter
& 1.180$\times$ \\
\bottomrule
\end{tabularx}
\end{table}

Table~\ref{tab:small drafter-choice} compares the two small drafter choices under the same middle--target pairs.
Replacing the neural small drafter with the n-gram drafter consistently improves speedup in all three configurations.
On average, the speedup increases from $0.865\times$ to $1.180\times$, corresponding to a relative gain of $36.4\%$.
In the two Qwen settings, using Qwen3-0.6B as the small drafter even leads to speedups below $1\times$, meaning that this configuration becomes slower than target-only autoregressive decoding.

These results support our main design choice.
In the shared-tree setting, using two neural drafters does not necessarily improve efficiency, because the forward cost of the neural small drafter can offset the benefit of speculative decoding.
The n-gram drafter provides a cheaper small drafter endpoint, allowing TreeGraft to reserve neural forwarding cost for the selectively invoked middle drafter.

\subsection{Stability of Online Evaluation}
\label{app:online-evaluation-stability}

To check whether the main online results are sensitive to wall-clock measurement noise, we repeat the online evaluation of the scheduler used in the main experiments. 
The two runs use the same scheduler parameters, identified tree signals, benchmarks, evaluation inputs, and decoding hyperparameters. 
The only difference is that the wall-clock measurements are collected from two independent online evaluation runs. 
We denote the results used in Table~\ref{tab:main_results} as Run 1, and the repeated evaluation as Run 2.

We evaluate two representative target--middle pairs: Qwen3-32B/Qwen3-0.6B and LLaMA 3.3-70B-Instruct/LLaMA 3.2-1B-Instruct. 
The former represents a Qwen setting where the benefit depends more on selective middle drafter calls, while the latter represents a LLaMA setting where the middle drafter is substantially smaller than the target model. 
Table~\ref{tab:online-evaluation-stability} reports the average speedup over six benchmarks.

\begin{table}[H]
\centering
\caption{Stability of repeated online evaluation. Run 1 denotes the result used in the main table, and Run 2 denotes an independent repeated evaluation with the same scheduler and decoding settings. Values are average speedups over six benchmarks.}
\label{tab:online-evaluation-stability}
\small
\begin{tabular}{lccc}
\toprule
Target--middle pair & Run 1 & Run 2 & Abs. diff. \\
\midrule
Qwen3-32B / Qwen3-0.6B
& 1.470$\times$ & 1.473$\times$ & 0.003$\times$ \\
LLaMA 3.3-70B-Instruct / LLaMA 3.2-1B-Instruct
& 2.297$\times$ & 2.289$\times$ & 0.008$\times$ \\
\midrule
Average
& 1.884$\times$ & 1.881$\times$ & 0.003$\times$ \\
\bottomrule
\end{tabular}
\end{table}

The repeated evaluation produces nearly identical results. 
Across the two representative target--middle pairs, the average speedup changes from $1.884\times$ to $1.881\times$, with an absolute difference of only $0.003\times$. 
This suggests that the main online results are stable under repeated wall-clock evaluation, and that the conclusions in Section~\ref{sec:experiments} do not depend on a single timing run.

\subsection{Complete Main Results and Endpoint Synergy}
\label{app:full-main-results}

Table~\ref{tab:full_main_results} gives the complete per-task speedups underlying the compact main-text table.

\fullmainresultstable

\subsection{Scheduler Transfer Across Temperature, Width, Models, and Hardware}
\label{app:sampling-transfer}

We test no-refit scheduler transfer across four deployment changes: sampling temperature ($T=0\rightarrow1$), candidate-tree width ($W=10\rightarrow5$), model family, and hardware (A100$\rightarrow$H20). For temperature transfer, we directly apply the scheduler fitted at $T=0$ to $T=1$ across all 60 main-table settings. As shown in Table~\ref{tab:temperature-transfer}, TreeGraft reaches 1.55$\times$, outperforming All-Small (1.34$\times$) and All-Mid (1.01$\times$). As a control, we recollect call/skip trajectories and refit the scheduler at $T=1$, obtaining a comparable speedup of 1.56$\times$. Thus, the no-refit result shows that the scheduler fitted at $T=0$ can transfer to non-greedy sampling at $T=1$.

\begin{table}[H]
\centering
\caption{Average speedup under sampling decoding with $T=1$.}
\label{tab:temperature-transfer}
\small
\begin{tabular}{lccc}
\toprule
Method & Setting & Average speedup & Improvement over better fixed endpoint \\
\midrule
All-Small ($T=1$) & --- & 1.34$\times$ & --- \\
All-Mid ($T=1$) & --- & 1.01$\times$ & --- \\
TreeGraft ($T=1$) & Refit & \textbf{1.56$\times$} & +16.4\% \\
TreeGraft ($T=1$) & No refit & \textbf{1.55$\times$} & +15.7\% \\
\bottomrule
\end{tabular}
\end{table}

We directly apply the scheduler fitted at $W=10$ to $W=5$ across all 60 main-table settings, with no refitting. As shown in Table~\ref{tab:width-transfer}, at $W=5$, TreeGraft reaches 1.50$\times$ versus 1.32$\times$ and 1.24$\times$ for the endpoints, a 13.6\% improvement over the better fixed endpoint. This shows that the scheduler can transfer to some extent under a change in candidate-tree width.

\begin{table}[H]
\centering
\caption{Average speedup in no-refit scheduler transfer under a change in candidate-tree width.}
\label{tab:width-transfer}
\small
\begin{tabular}{lcccc}
\toprule
Tree width & TreeGraft & All-Small & All-Mid & Improvement over better fixed endpoint \\
\midrule
$W=10$ & \textbf{1.60$\times$} & 1.32$\times$ & 1.39$\times$ & +15.1\% \\
$W=5$ & \textbf{1.50$\times$} & 1.32$\times$ & 1.24$\times$ & +13.6\% \\
\bottomrule
\end{tabular}
\end{table}

Table~\ref{tab:main_results} already provides evidence of scheduler transfer to four held-out model pairs from seen families and to MT-Bench, which was excluded from fitting. We further test both cross-family directions (30 settings each) and A100$\rightarrow$H20 transfer (all 60 settings). For each cross-family direction, the scheduler is fitted using the source-family fitting pairs specified in Appendix~\ref{app:fitting_data} and then evaluated on the other family without refitting. For A100$\rightarrow$H20, the scheduler fitted on A100 is directly evaluated on H20 without refitting. As shown in Table~\ref{tab:family-hardware-transfer}, TreeGraft improves over the better fixed endpoint by 14.6\%, 6.5\%, and 32.1\%, respectively. These results indicate that the scheduler can transfer to some extent across model pairs, model families, workload, and hardware without refitting.

\begin{table}[H]
\centering
\caption{Average speedup in transfer evaluation across model families and hardware.}
\label{tab:family-hardware-transfer}
\small
\setlength{\tabcolsep}{3pt}
\begin{tabular}{lcccc}
\toprule
Deployment change & TreeGraft & All-Small & All-Mid & Improvement over better fixed endpoint \\
\midrule
Qwen fitting $\rightarrow$ LLaMA testing & \textbf{1.65$\times$} & 1.44$\times$ & 1.32$\times$ & +14.6\% \\
LLaMA fitting $\rightarrow$ Qwen testing & \textbf{1.32$\times$} & 1.24$\times$ & 0.95$\times$ & +6.5\% \\
A100 $\rightarrow$ H20 & \textbf{1.85$\times$} & 1.40$\times$ & 1.15$\times$ & +32.1\% \\
\bottomrule
\end{tabular}
\end{table}

\subsection{Scheduler Feature Ablations}
\label{app:scheduler-feature-ablation}

We separately remove runtime context, scheduling history, and tree signals and refit the scheduler for each ablation. The comparison covers all 60 settings in the main table. As shown in Table~\ref{tab:scheduler-feature-ablation}, relative to the corresponding ablated variants, the full scheduler improves average speedup by 9.6\%, 10.3\%, and 7.4\%, respectively. The largest observed contributions come from scheduling history and runtime context, while tree signals also contribute to performance.

\begin{table}[H]
\centering
\caption{Ablation of scheduler features.}
\label{tab:scheduler-feature-ablation}
\small
\begin{tabular}{lcc}
\toprule
Scheduler & Average speedup & Full-scheduler speedup improvement \\
\midrule
Without runtime-context features & 1.46$\times$ & +9.6\% \\
Without scheduling-history features & 1.45$\times$ & +10.3\% \\
Without tree-signal features & 1.49$\times$ & +7.4\% \\
Full learned scheduler & \textbf{1.60$\times$} & -- \\
\bottomrule
\end{tabular}
\end{table}

\subsection{Comparison with Heuristic Schedulers}
\label{app:heuristic-schedulers}

We evaluate four heuristic schedulers over all 60 main-table settings: early-call, entropy/confidence, candidate-score-gap, and latency-normalized. For each heuristic, we report the single global parameter from a predefined grid that maximizes average speedup. This provides an optimistic upper bound on each heuristic's performance within the predefined grid; the selected value appears in the method name.

\begin{table}[H]
\centering
\caption{Learned scheduler vs. heuristic baselines.}
\label{tab:heuristic-schedulers}
\small
\setlength{\tabcolsep}{3pt}
\begin{tabular}{lcc}
\toprule
Method & Average speedup & Learned-scheduler speedup improvement \\
\midrule
Always-Early (call in the first 3 drafting steps) & 1.48$\times$ & +8.1\% \\
Entropy/Confidence ($\tau=0.7$) & 1.25$\times$ & +28.0\% \\
Candidate-Score-Gap ($\tau=0.7$) & 1.46$\times$ & +9.6\% \\
Latency-Normalized ($\tau=1.0$) & 1.50$\times$ & +6.7\% \\
Learned scheduler & \textbf{1.60$\times$} & -- \\
\bottomrule
\end{tabular}
\end{table}

As Table~\ref{tab:heuristic-schedulers} shows, the strongest heuristic, Latency-Normalized, achieves 1.50$\times$ but remains below the learned scheduler's 1.60$\times$. This gap is consistent with the fact that each heuristic captures only one aspect of the scheduling state, whereas the learned scheduler combines runtime context, tree signals, and observed scheduling history to estimate the middle drafter's marginal benefit and latency cost.

\section{Existing Assets and Licenses}
\label{app:assets}
We use publicly available pretrained checkpoints and benchmarks, including LLaMA models, Qwen3 models, GSM8K, Alpaca, NQ, HumanEval, CNN/DM, and MT-Bench. We cite the original sources for all models and datasets and use them only for research evaluation. The LLaMA checkpoints are used under the Meta LLaMA license and acceptable-use policy, Qwen3 checkpoints are released under the Apache License 2.0, GSM8K and HumanEval are released under the MIT License, Alpaca is released for research use under CC BY-NC 4.0, and Natural Questions is released under Apache License 2.0. For CNN/DailyMail and MT-Bench, we follow the terms and licenses specified by their
official dataset/model-card or repository distributions. For datasets whose redistributed versions have additional terms, we follow the terms provided by the corresponding dataset hosts.

%% file: sections/checklist.tex
\section*{NeurIPS Paper Checklist}

\begin{enumerate}

\item {\bf Claims}
    \item[] Question: Do the main claims made in the abstract and introduction accurately reflect the paper's contributions and scope?
    \item[] Answer: \answerYes{}
    \item[] Justification: The abstract and introduction state the main scope of TreeGraft as a training-free multi-drafter framework for shared-tree speculative decoding, and the reported claims are supported by the method design in Sections~\ref{sec:grafting}--\ref{sec:online_scheduling} and the empirical results in Section~\ref{sec:experiments}.

\item {\bf Limitations}
    \item[] Question: Does the paper discuss the limitations of the work performed by the authors?
    \item[] Answer: \answerYes{}
    \item[] Justification: The paper includes a separate Limitations section discussing the training-free scope of TreeGraft and the use of an n-gram drafter as the low-cost small drafter. These limitations clarify that TreeGraft is orthogonal to training-based draft-source improvement and that the lookup-based small drafter is one practical instantiation rather than an exhaustive choice. (See Appendix~\ref{app:limitations})

\item {\bf Theory assumptions and proofs}
    \item[] Question: For each theoretical result, does the paper provide the full set of assumptions and a complete (and correct) proof?
    \item[] Answer: \answerNA{}
    \item[] Justification: The paper does not present formal theoretical results, theorems, or proofs. The mathematical expressions are used to define the algorithmic components and scheduling objective.

    \item {\bf Experimental result reproducibility}
    \item[] Question: Does the paper fully disclose all the information needed to reproduce the main experimental results of the paper to the extent that it affects the main claims and/or conclusions of the paper (regardless of whether the code and data are provided or not)?
    \item[] Answer: \answerYes{}
    \item[] Justification: The paper provides the model pairs, datasets, baselines, metrics, tree-construction hyperparameters, offline fitting procedure, scheduler training details, and online evaluation protocol in Section~\ref{sec:experiments} and Appendices~\ref{app:online_scheduling}--\ref{app:online_eval_setting_details}.

\item {\bf Open access to data and code}
    \item[] Question: Does the paper provide open access to the data and code, with sufficient instructions to faithfully reproduce the main experimental results, as described in supplemental material?
    \item[] Answer: \answerYes{}
    \item[] Justification: We provide a code repository at \url{https://github.com/fjm9933/TreeGraft}. The experiments use public benchmarks and publicly available pretrained checkpoints. Together with the algorithmic details, hyperparameters, tree-construction settings, scheduler descriptions, and evaluation protocols provided in the paper and supplemental material, the released code supports reproduction of the main experimental results.

\item {\bf Experimental setting/details}
    \item[] Question: Does the paper specify all the training and test details (e.g., data splits, hyperparameters, how they were chosen, type of optimizer) necessary to understand the results?
    \item[] Answer: \answerYes{}
    \item[] Justification: Section~\ref{sec:experiments} specifies the evaluated model pairs, datasets, baselines, metrics, and decoding hyperparameters. Appendices~\ref{app:online_scheduling} and~\ref{app:online_eval_setting_details} further describe fitting data collection, online states, scheduler training, evaluation subsets, and runtime settings.

\item {\bf Experiment statistical significance}
    \item[] Question: Does the paper report error bars suitably and correctly defined or other appropriate information about the statistical significance of the experiments?
    \item[] Answer: \answerYes{}
    \item[] Justification: Appendix~\ref{app:online-evaluation-stability} reports repeated-run stability statistics for the
    timing-based online speedup results under the same scheduler, inputs, and decoding
    hyperparameters. We use this as variability information.

\item {\bf Experiments compute resources}
    \item[] Question: For each experiment, does the paper provide sufficient information on the computer resources (type of compute workers, memory, time of execution) needed to reproduce the experiments?
    \item[] Answer: \answerYes{}
    \item[] Justification: Appendix~\ref{app:online_eval_setting_details} reports the hardware and runtime configuration used for online evaluation, including the use of two NVIDIA A100 GPUs, batch size 1. Appendix~\ref{app:runtime} also reports representative per-call latency measurements for target and drafter models.
    
\item {\bf Code of ethics}
    \item[] Question: Does the research conducted in the paper conform, in every respect, with the NeurIPS Code of Ethics \url{https://neurips.cc/public/EthicsGuidelines}?
    \item[] Answer: \answerYes{}
    \item[] Justification: The research uses public language-model checkpoints and public text benchmarks, and does not involve human-subject experiments, private data, or sensitive personal information. We are not aware of any deviation from the NeurIPS Code of Ethics.

\item {\bf Broader impacts}
    \item[] Question: Does the paper discuss both potential positive societal impacts and negative societal impacts of the work performed?
    \item[] Answer: \answerYes{}
    \item[] Justification: The work may reduce the computational cost and latency of LLM inference, which can improve accessibility and energy efficiency. At the same time, faster inference could also lower the cost of generating harmful or misleading text, so responsible deployment should follow the safeguards of the underlying language models.
    
\item {\bf Safeguards}
    \item[] Question: Does the paper describe safeguards that have been put in place for responsible release of data or models that have a high risk for misuse (e.g., pre-trained language models, image generators, or scraped datasets)?
    \item[] Answer: \answerNA{}
    \item[] Justification: The paper does not release a new pretrained language model, image generator, scraped dataset, or other high-risk asset. TreeGraft is an inference acceleration framework applied to existing models.

\item {\bf Licenses for existing assets}
    \item[] Question: Are the creators or original owners of assets (e.g., code, data, models), used in the paper, properly credited and are the license and terms of use explicitly mentioned and properly respected?
    \item[] Answer: \answerYes{}
    \item[] Justification: The paper cites the original sources for all pretrained models and benchmarks used in the experiments and includes an existing-assets statement in Appendix~\ref{app:assets}. We use these assets for research evaluation and follow their corresponding licenses and terms of use.

\item {\bf New assets}
    \item[] Question: Are new assets introduced in the paper well documented and is the documentation provided alongside the assets?
    \item[] Answer: \answerNA{}
    \item[] Justification: The paper does not introduce or release a new dataset, pretrained model, or benchmark asset. The contribution is an algorithmic framework and evaluation protocol for inference acceleration.

\item {\bf Crowdsourcing and research with human subjects}
    \item[] Question: For crowdsourcing experiments and research with human subjects, does the paper include the full text of instructions given to participants and screenshots, if applicable, as well as details about compensation (if any)? 
    \item[] Answer: \answerNA{}
    \item[] Justification: The paper does not involve crowdsourcing experiments or research with human subjects.

\item {\bf Institutional review board (IRB) approvals or equivalent for research with human subjects}
    \item[] Question: Does the paper describe potential risks incurred by study participants, whether such risks were disclosed to the subjects, and whether Institutional Review Board (IRB) approvals (or an equivalent approval/review based on the requirements of your country or institution) were obtained?
    \item[] Answer: \answerNA{}
    \item[] Justification: The paper does not involve human-subject research, so IRB approval or equivalent review is not applicable.

\item {\bf Declaration of LLM usage}
    \item[] Question: Does the paper describe the usage of LLMs if it is an important, original, or non-standard component of the core methods in this research? Note that if the LLM is used only for writing, editing, or formatting purposes and does \emph{not} impact the core methodology, scientific rigor, or originality of the research, declaration is not required.
    \item[] Answer: \answerNA{}
    \item[] Justification: The core method development, experimental design, and scientific conclusions of this work do not rely on LLMs as an important, original, or non-standard research component. Any use of LLM-based tools, if any, was limited to writing, editing, or formatting assistance and did not affect the methodology, experiments, or results.

\end{enumerate}